# Simplified Swarm Optimization for Surrogate-Assisted Reliability Design of Insulated-Gate Bipolar Transistor Power Modules Using an Open-Source Process Finite-Element Model†

Wei-Chang Yeh[1,*]

[1] Department of Industrial Engineering and Engineering Management, National Tsing Hua University, Hsinchu 300044, Taiwan

* Corresponding author: wcyeh@ie.nthu.edu.tw

**Abstract**

Process-induced warpage, ceramic stress and solder strain limit the reliability of insulated-gate bipolar transistor (IGBT) modules on direct-bonded copper (DBC) substrates. Surrogate-assisted design studies train regression models on finite-element analysis (FEA) databases, but rarely check the optimized designs against new FEA or report how surrogate error interacts with the optimizer. This paper builds and evaluates an open pipeline: an open-source process finite-element model, surrogates tuned by Simplified Swarm Optimization (SSO), multi-objective design search, FEA confirmation of selected designs and confirmation-driven infill. The model starts at the second reflow and reproduces measured warpage within 18.2%, 33.6% and 15.3% at the reflow, housing and molding stages without fitted parameters. On a balanced 60-design database, all stochastic tuners reach the same test accuracy, outperform the published grid on cross-validated $R^2$ for every output but generalize better only for warpage; the cross-validated ranking of tuners does not transfer to the test set. With equal result reporting, multi-objective SSO and the non-dominated sorting genetic algorithm II give comparable Pareto fronts; a corrected multi-objective particle swarm optimizer trails both. FEA confirmation shows that warpage predictions hold (mean absolute error 0.5%), whereas at the design-space bounds reached by the optimizers the ceramic-stress surrogate is optimistic by up to 26%. Two confirmation-driven infill rounds reduce this error to 1–10% and halve the out-of-sample error; no confirmed design improves on the database in ceramic stress. Ceramic-stress results are indicative, as the metric is mesh-sensitive at production resolution. The model, database, scripts and pre-registered and post-registration results are released.

† Submitted to Engineering Applications of Artificial Intelligence.

## 1. Introduction

Power modules built on IGBT chips and DBC substrates carry the switching load of electric vehicles and renewable-energy converters, and their service life is limited by thermo-mechanical damage. The mismatch in coefficient of thermal expansion (CTE) between silicon, solder, copper, ceramic and epoxy generates process-induced warpage during reflow and molding, tensile stress in the brittle ceramic, and accumulated plastic strain in the solder joints. Each of these responses depends on the thicknesses of the copper and ceramic layers and on the ceramic material, which makes the substrate stack a natural design variable.

FEA can predict these responses, but a process simulation with viscoplastic solder and cure-shrinking epoxy is expensive; Lee, Tsai and Chuang report 604,800 s per FEA case for such a module [1]. To reduce this cost, recent studies train machine-learning surrogates on FEA databases. Lee et al. [1] validated a commercial process model against measured warpage at four process stages, generated 60 FEA cases over layer thicknesses and four ceramic materials, and trained SVR models for warpage (test $R^2$ = 0.9891), ceramic maximum principal stress (0.9908) and solder accumulated plastic strain (0.9619).

Two steps remain open in that framework. First, the SVR hyperparameters were chosen by a grid search of 10 × 10 × 10 × 2 = 2,000 combinations, and several reported optima lie outside the stated grid; a continuous search can reach settings that a grid cannot. Second, the study names multi-objective optimization with genetic algorithms or particle swarm optimization as the final step of its design strategy but does not carry it out. Its design space is also discrete (3 × 5 × 3 × 4 = 180 combinations), which a surrogate can relax to continuous thicknesses.

This paper builds both steps into an open pipeline and examines what happens when its output is checked by FEA. SSO [2], a population-based metaheuristic whose update rule sets each variable to the global best, the personal best, the current value or a random value with fixed probabilities, is used to tune the surrogates and, in its multi-objective form, to search the design space. The contributions are fourfold. First, an open, documented process FE model of the reference module, validated stage by stage without fitted parameters. Second, a pre-registered comparison of surrogate tuners and of multi-objective optimizers at equal budgets, with every result reported against its pre-registered protocol and every later correction labeled. Third, FEA confirmation of the selected designs followed by confirmation-driven infill, which quantifies how the optimizers exploit surrogate error at the design-space bounds and how much of that error two infill rounds remove. Fourth, the FE model, the databases and all scripts are released.

Section 2 lists the acronyms, notations, nomenclatures and assumptions. Section 3 reviews related work and states the two problems solved. Section 4 describes the FE model and its validation, and Section 5 presents the SSO framework. Section 6 defines the experimental design, Section 7 reports the results and Section 8 discusses them. Section 9 concludes.

## 2. Acronyms, Notations, Nomenclatures, and Assumptions

This section collects the abbreviations, symbols and terms used throughout the paper, and the modeling assumptions on which all results depend.

**Acronyms.** AlN, aluminum nitride; CTE, coefficient of thermal expansion; DBC, direct bonded copper; EMC, epoxy molding compound; FE and FEA, finite element and finite-element analysis; GA, genetic algorithm; IGBT, insulated-gate bipolar transistor; IQR, interquartile range; MAPE, mean absolute percentage error; MOPSO, multi-objective particle swarm optimization; NSGA-II, non-dominated sorting genetic algorithm II; PSO, particle swarm optimization; RBF, radial basis function; RMSE, root-mean-square error; SAC305, Sn-3.0Ag-0.5Cu lead-free solder; SSO, simplified swarm optimization; SVR, support vector regression; TPE, tree-structured Parzen estimator; ZTA, zirconia-toughened alumina.

**Notations.** The symbols are listed in Table 1. Scalars are italic, vectors bold and sets calligraphic, and a hat denotes an SVR prediction.

Table 1. Notations.

| Symbol | Meaning |
|---|---|
| $i$, $j$, $t$ | index of a solution, index of a variable and iteration number |
| $x_{ij}^{t}$ | value of variable $j$ of solution $i$ at iteration $t$ |
| $g_j$, $p_{ij}$ | variable $j$ of the global best solution and of the personal best of solution $i$ |
| $x$ | random value drawn from the range of variable $j$ |
| $\rho$ | random number drawn from $U(0,1)$ |
| $C_g$, $C_p$, $C_w$ | SSO thresholds, $0<C_g<C_p<C_w<1$ |
| $t_{\mathrm{tc}}$, $t_{\mathrm{c}}$, $t_{\mathrm{bc}}$ | top-copper, ceramic and bottom-copper thickness (mm) |
| $m$, $M$ | ceramic material and the set $\{\mathrm{AlN}, \mathrm{Al_2O_3}, \mathrm{ZTA}, \mathrm{Si_3N_4}\}$ |
| $d$, $D$ | design vector and design space |
| $F(d)$ | vector of the three predicted objectives |
| $w$, Ω | out-of-plane displacement and warpage measurement region |
| $W$ | warpage at the end of molding (μm) |
| $\sigma_{\mathrm{c}}$ | maximum over the three process stages of the signed, region-averaged maximum principal stress in the ceramic (MPa) |
| $\varepsilon_{\mathrm{p}}$ | region-averaged accumulated plastic strain in the solder at the end of molding |

| Symbol | Meaning |
|---|---|
| $\widehat{W}$, $\hat{\sigma}_{\mathrm{c}}$, $\hat{\varepsilon}_{\mathrm{p}}$ | SVR predictions of $W$, $\sigma_{\mathrm{c}}$ and $\varepsilon_{\mathrm{p}}$ |
| $y$ | one of the three outputs $W$, $\sigma_{\mathrm{c}}$, $\varepsilon_{\mathrm{p}}$ |
| $C$, $\varepsilon$, $\gamma$ | SVR penalty, tube width and RBF kernel width |
| $\theta$, $\Theta$ | SVR hyperparameter vector and its bounds |
| $R^2$, $R^2_{\mathrm{CV}}$ | coefficient of determination on the test cases and its five-fold cross-validated value on the training cases |
| $E$, $v$, $T_{\mathrm{g}}$ | Young's modulus, Poisson's ratio and glass transition temperature |

**Nomenclatures.** A design is one combination of top-copper, ceramic and bottom-copper thickness and ceramic material. A process stage is one of the four measured states of the module in the reference study: after the first reflow, after the second reflow on the baseplate, after the housing bake and after molding. The FEA-v5.3 model reproduces the last three, starting from the second reflow. An FEA case is one full process simulation of one design, from the second reflow to the end of molding, and FEA-v5.3 is the frozen production FE model used for every FEA case in this paper. A model fit is one SVR training with a given hyperparameter setting, including its five-fold cross-validation, and is the unit of the tuning budget. A surrogate evaluation is one prediction of $F(d)$ by the three SVR models and is the unit of the design-search budget. The Pareto front is the set of designs not dominated in all three objectives by any other design found, and its knee point is the design with the best balance of the three normalized objectives.

**Assumptions.** The FE model and the optimization rest on the following assumptions, which are revisited in Section 8.

A1. Dimensions, materials and process conditions follow the reference study [1]. Quantities it does not report are assumed: the housing is a ring frame with 3.5 mm side walls, the chip layout is symmetric, the warpage measurement region is $|x| \leq 56.45$ mm, and the pre-bend is a paraboloid scaled to +599.6 µm.

A2. Silicon and the ceramics are linear elastic, copper is elasto-plastic, and the housing is anisotropic linear elastic. The EMC is linear elastic with a 0.192% volumetric cure shrinkage applied during the 150 °C bake; its viscoelasticity is neglected.

A3. The solder is stress-free at 217 °C, and its steady-state Anand flow rule is represented by temperature-dependent Norton creep; transient hardening is neglected.

A4. The simulation starts at the second reflow; the first (die-attach) reflow is not simulated, and both solder layers are born strain-free at 217 °C on cooling. Parts are activated by element birth in the process order, parts not yet born are carried by soft placeholder elements, and rigid-body motion is removed by a 3-2-1 constraint at the baseplate corners.

A5. Region-averaged stress and strain replace peak values at geometric singularities, and the mesh bias of these outputs is taken to be uniform across designs (checked post-registration on four designs, Supplementary Table S7: not confirmed for the ceramic stress, see Section 4.2).

A6. Layer thicknesses are continuous within the bounds of the reference grid, the ceramic material is categorical, and the surrogates are used only inside these bounds.

A7. All tuners and optimizers use the same FE model, data, budgets and hardware, so their comparison does not depend on the absolute accuracy of the FE model.

## 3. Related Work and the Proposed Problem

### 3.1. Related Work

**FEA-based machine learning for packaging reliability.** Surrogates trained on FEA data now predict warpage, stress and solder-joint life across packaging technologies. An artificial neural network trained on a finite-element database predicts the reliability of wafer-level packages [3], and SVR predicts their solder-joint life [4], the closest precedent for the surrogates of this paper. Lee et al. use SVR for IGBT-DBC modules [1]; the same group applies SVR to warpage of redistribution-layer glass substrates, reporting a test $R^2$ of 0.9995 [5], and a regression surrogate to die shift in panel-level packaging [6]. Deep networks trained on finite-element warpage cover fan-out wafer-level packages [7] and memory packages [8], and a point-cloud model predicts solder-joint reliability from three-dimensional FEA data [9]. Operator-learning and graph models extend prediction from scalars to fields: DeepONet for package warpage [10], a physics-aware graph neural network for multi-die warpage [11] and a mode-aware hybrid model for full-field warpage of fan-out panel-level packages [12]. Fast physics solvers, such as the two-dimensional laminate-theory FEM of Lo, Liu and Chang [13], and reduced-order models with neural-network field reconstruction for the co-simulation of power modules [14] reduce the cost of each evaluation, and neural networks also predict the thermal resistance of heterogeneous integrated chips [15]. These works focus on prediction accuracy; design optimization on top of the surrogate is carried out in only a few of them, for example by NSGA-II on random-forest and gradient-boosting surrogates of 2.5D flip-chip packages [16] and by adaptive Bayesian optimization of a Gaussian-process surrogate of the solder creep strain [17].

**Process modeling of molded modules.** Failure mechanisms of power modules, among them solder fatigue, are reviewed by Ciappa [18]. Warpage of molded modules depends on the epoxy molding compound (EMC), whose properties evolve during cure [19] and whose critical properties for warpage prediction have been reviewed [20]; Lee and Lin model its viscoelasticity and cure shrinkage with a two-domain modified Tait formulation [21]. Process-sequence simulations reproduce the warpage of intelligent power modules [22], the process-induced loading of epoxy-encapsulated high-power modules [23] and the residual stress and permanent warpage of resin-molded SiC power modules [24]. Solder creep is commonly described by the Anand viscoplastic model [25,26], which has been applied to solder alloys [27] and compared with other constitutive models for solders [28]; calibrated SAC305 models have been validated against in-situ thermo-mechanical fatigue measurements [29], and the properties and fatigue life of SnAgCu joints have been evaluated for power-module applications [30]. The Anand model and cure-dependent EMC models are available in commercial solvers but not in most open-source codes, which motivates the simplifications documented in Section 4. Open-source finite-element codes have been used for thermal problems in electronics, for example parallel FEniCSx models of heat sinks [31], but this review found no peer-reviewed open-source process simulation of a molded power module, the gap that the model of Section 4 addresses.

**Surrogate-assisted optimization and hyperparameter search.** When each evaluation is expensive, a surrogate is fitted to sampled designs and optimized instead, and the optimum is then confirmed by the true model [32,33]. Efficient global optimization chooses each new sample by its expected improvement [34], and a taxonomy of response-surface methods shows how an inaccurate surrogate can mislead the search [35]. Surrogate-assisted evolutionary computation manages the surrogate during the search [36], and adaptive sampling adds samples where the metamodel is uncertain or the designs are promising [37], the role of the confirmation-driven infill of this paper. In engineering design, SVR surrogates [38] and least-squares SVR surrogates tuned by particle swarm optimization [39] have been optimized and their optima checked by finite-element analysis. The accuracy of an $\varepsilon$-SVR [40] depends strongly on the penalty $C$, the tube width $\varepsilon$ and the radial-basis-function (RBF) width $\gamma$, and grid search, random search [41], Bayesian optimization such as the tree-structured Parzen estimator of Optuna [42] and metaheuristics are all used to set them. On small data sets, over-fitting of the model-selection criterion can make a cross-validated ranking of tuned models unreliable [43].

**Multi-objective metaheuristics and their assessment.** GA, PSO [44] and their multi-objective forms, NSGA-II [45] and MOPSO [46], are the standard baselines for engineering design, together with decomposition-based methods such as MOEA/D, the multi-objective evolutionary algorithm based on decomposition [47]. Multi-objective particle swarm optimizers have been surveyed by Reyes-Sierra and Coello Coello [48], and the speed-constrained multi-objective PSO (SMPSO) limits the particle velocity to stabilize the search [49]; pymoo provides a widely used implementation of NSGA-II [50]. Approximations of the Pareto front are compared with quality indicators such as the hypervolume, whose properties are analyzed by Zitzler et al. [51], and stochastic optimizers are compared over repeated runs with non-parametric tests, the Friedman test with post-hoc procedures [52] and the Wilcoxon signed-rank test with Holm correction [53].

**Simplified swarm optimization.** SSO [2] replaces the velocity update of PSO with a stepwise selection rule that has few parameters and treats discrete and continuous variables alike. It has been applied to reliability redundancy allocation, including a bi-objective variant [54], to neural-network training [55] and to hyperparameter tuning of convolutional neural networks [56,57]. Its multi-objective forms include a weighting scheme for feature selection with support vector machines [58], non-dominated sorting variants for multi-stage capacitated facility location [59] and omni-channel pollution routing [60], and a two-objective SSO for cloud computing compared with NSGA-II and MOPSO [61]; a recent study in this journal applies SSO to cutting optimization problems [62]. Its use for FEA-surrogate design of power-module packaging is the subject of this paper.

### 3.2. Proposed Problem

The review identifies two open problems in the reference framework [1], which this paper states formally and solves with SSO. Figure 1 shows the modeled module, the design variables with their ranges, where the three outputs are evaluated and the process sequence. Both are defined on the FE database of Section 6 and on the three outputs $W$, $\sigma_c$ and $\varepsilon_p$ of Section 4.2.

**Problem P1: surrogate tuning.** For each output $y \in \{W, \sigma_c, \varepsilon_p\}$, an $\varepsilon$-SVR with an RBF kernel is tuned by

$$\max_{\theta \in \Theta} R^2_{CV}(\theta; y), \theta = (\log_{10} C, \log_{10} \varepsilon, \log_{10} \gamma), \quad (1)$$

where $\Theta$ is a box of hyperparameter bounds $\log_{10} C \in [-1,4]$, $\log_{10} \varepsilon \in [-3,0]$ and $\log_{10} \gamma \in [-3,1]$ on $z$-scored data (the upper $C$ bound was widened from the pre-registered $10^3$ before launch, after a smoke test placed the optimum at $C = 759 - 979$; a documented deviation) and $R^2_{\mathrm{CV}}$ is the five-fold cross-validated coefficient of determination on the 48 training cases. The 12 test cases are not used in P1. The reference study solved a discrete version of P1 with a grid of 2,000 combinations [1]; here P1 is continuous, and its budget is counted in model fits.

**Problem P2: multi-objective design.** With the design vector $d = (t_{\mathrm{tc}}, t_{\mathrm{c}}, t_{\mathrm{bc}}, m)$, the design problem is

$$\min_{d \in D} F(d) = \left(\widehat{W}(d), \hat{\sigma}_{\mathrm{c}}(d), \hat{\varepsilon}_{\mathrm{p}}(d)\right), \tag{2}$$

where $D = [0.1, 0.3] \times [0.25, 0.635] \times [0.1, 0.3] \times M$, with thicknesses in mm, mixes three continuous variables with the categorical ceramic material, and the hats denote the SVR models obtained from P1. The reference study named this step but did not solve it, and its design space was the discrete grid of 180 combinations [1].

**Role of the FE model.** One FEA case takes about 2.9 h, so the FE model is used only to build the training data and to confirm the final designs. A solution of P2 counts as a design improvement only after its FEA confirmation (Section 5.3).

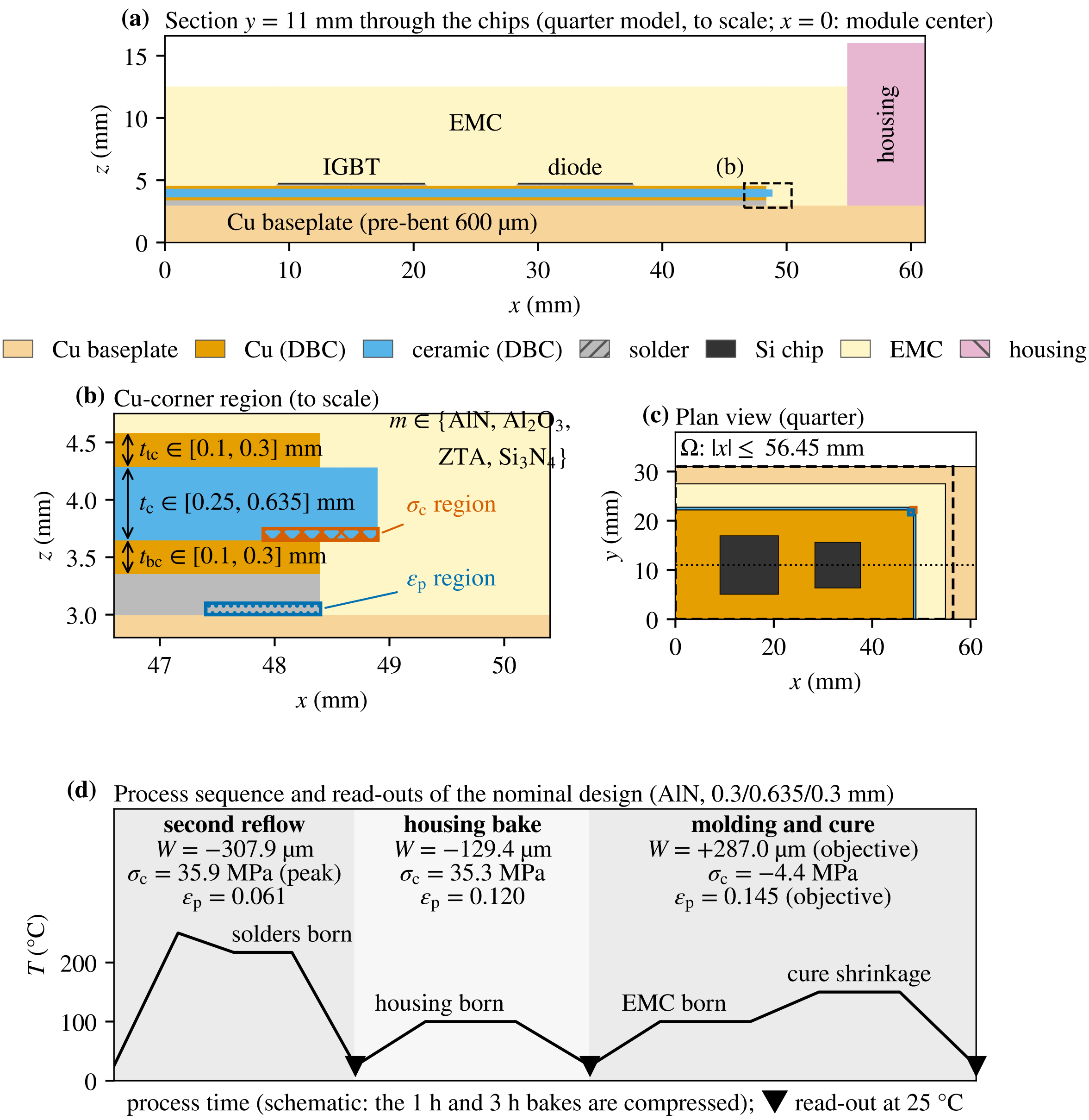

Figure 1. Problem setting. (a) Section $y = 11$ mm through the chips of the quarter model of the nominal module (AlN; top copper, ceramic and bottom copper 0.3/0.635/0.3 mm), drawn to scale from the FE model geometry; $x = 0$ is the module center. (b) Cu-corner region, to scale: the design variables $t_{tc}$, $t_c$ and $t_{bc}$ (top-copper, ceramic and bottom-copper thickness) with their ranges and the ceramic material $m$ of Problem P2, and the regions in which the ceramic stress $\sigma_c$ (maximum principal stress averaged over 1.0 × 1.0 × 0.1 mm at the Cu corner, maximum over the stages; mesh-sensitive, Section 4.2) and the solder strain $\varepsilon_p$ (accumulated plastic strain averaged over 1.0 × 1.0 × 0.1 mm at the solder corner, end of molding) are evaluated, drawn at the faces that govern for the nominal design. (c) Plan view of the quarter model: the warpage $W$ is the range of the baseplate-bottom deflection inside Ω ($|x| \le 56.45$ mm); the dotted line is the section of (a), the squares are the $\sigma_c$ and $\varepsilon_p$ regions. (d) Process

sequence of the FE model (temperatures to scale, durations schematic) and the read-outs of the nominal design at 25 °C after each stage: $\sigma_c$ peaks after the second reflow, as in 58 of the 60 database designs, while $W$ and $\varepsilon_p$ are taken at the end of molding.

## 4. Open-Source Process Finite-Element Model

### 4.1. Geometry, Materials and Process Sequence

The reference module of Lee et al. [1] is rebuilt in CalculiX [63] with meshes generated by gmsh [64], driven from Python so that each design is evaluated by a single function call. Every value in this section comes from the reference study unless stated as an assumption, and all assumptions are listed in the released assumption log. The component dimensions are given in Table 2.

Table 2. Component dimensions of the modeled IGBT-DBC module (from [1] unless noted).

| Component | Thickness (mm) | Length × width (mm) |
|---|---|---|
| IGBT chip | 0.14 | 11.9 × 11.9 |
| Diode chip | 0.14 | 9.3 × 9.3 |
| Solder under IGBT and diode | 0.1 | 12.2 × 12.2 and 9.6 × 9.6 |
| Solder, DBC to baseplate | 0.35 | 96.8 × 44.4 |
| Ceramic (AlN, nominal) | 0.635 | 97.8 × 45.5 |
| Copper layers, top and bottom | 0.3 | 96.8 × 44.4 |
| Copper baseplate, pre-bent 600 µm | 3 | 122.3 × 62 |
| Epoxy molding compound | 9.5 | 109.8 × 55 |
| Housing frame | 13 (height) | ring around the EMC footprint (assumed) |

The reference study gives the housing as a single 62 × 13 × 13 mm size. It is modeled here as a ring frame bonded to the baseplate around the EMC footprint with 3.5 mm side walls, an interpretation that proved decisive for the housing stage (Section 4.3). Chip positions, baseplate hole positions and the pre-bend shape are not reported; a symmetric chip layout, a warpage measurement region of $|x| \leq 56.45$ mm and a paraboloid pre-bend scaled to +599.6 µm are assumed. The material properties are listed in Table 3.

Table 3. Material properties ($E$: Young's modulus; $v$: Poisson's ratio; $T_g$: glass transition temperature).

| **Material** | $E$ **(GPa)** | $v$ | **CTE (ppm/°C)** | **Behavior** |
|---|---|---|---|---|
| Silicon | 169.5 | 0.28 | 3 | linear elastic |
| SAC305 solder | 14 at −40 °C to 4.85 at 125 °C | 0.4 | 22.5 | Anand-based creep (Section 4.2) |
| Copper | 118 | 0.35 | 17 | elasto-plastic, yield 173 MPa |
| AlN | 340 | 0.25 | 4.3 | linear elastic |
| $Al_2O_3$ | 340 | 0.22 | 7 | linear elastic |
| ZTA | 360 | 0.25 | 7 | linear elastic |
| $Si_3N_4$ | 320 | 0.28 | 2.5 | linear elastic |
| Housing composite | 20 | 0.4 | in plane 18; through thickness 27 below and 57 above $T_g$ = 120 °C | anisotropic |
| EMC | 20 | 0.4 | 15 (cured at 150 °C, below $T_g$ = 161 °C) | 0.192% volumetric cure shrinkage |

The process follows the four measured stages of the reference study. Solder, IGBT and diode are placed on the DBC at 25 °C and reflowed at 250 °C, above the 217 °C melting point of SAC305, then cooled. The DBC is placed on the copper baseplate pre-bent to 600 μm and reflowed a second time (measured warpage −376.4 μm). The housing is bonded at the baseplate edge and baked at 100 °C for 1 h (−194.8 μm). The EMC is injected into the frame and baked at 100 °C for 1 h and then at 150 °C for 3 h, during which the cure shrinkage is applied (+339.0 μm). Parts are activated step by step by element birth, the solder is stress-free at 217 °C, and the cooling rates of 1 K/s after reflow and 2 K/min after oven bakes are assumptions.

**Process start.** The simulation starts at the second reflow. The DBC with its chips and the pre-bent baseplate are stress-free at 25 °C with both solder layers absent; the assembly is heated to the 250 °C peak and cooled to 217 °C, where the die-attach and DBC-to-baseplate solder layers are born strain-free, and then cooled to 25 °C before the housing and molding stages. The first (die-attach) reflow is not simulated for three reasons. The first warpage measurement is taken after the second reflow; both solder layers remelt in the second reflow, so the first leaves little residual state; and an explicit first reflow leaves the DBC free-standing on placeholder elements, a stage that failed to converge in about 30% of the designs regardless of the linear solver (SPOOLES or PARDISO) or the support variant tried. Transferring a first-reflow state into a separate model is not possible in CalculiX 2.21 (a stress-only transfer failed a one-element verification test). Omitting the first reflow changes the baseline warpage errors by +4.2, +0.3 and −2.5 percentage points at the reflow, housing and molding stages (Section 4.3); this departure from the pre-set ±2-point acceptance rule is documented in the released assumption log.

**Placeholder elements.** Parts that are born later in the process (the solder layers, housing, adhesive and EMC) have coincident placeholder elements that keep the parts' nodes attached before birth and remain active throughout. They are soft (small-strain Young's modulus 1 MPa, Poisson's ratio 0.3) and have zero thermal expansion. In the production model FEA-v5.3 they follow a compressible Neo-Hookean law ($C_{10} = 0.192\,\mathrm{MPa}$, $D_1 = 2.4\,\mathrm{MPa}^{-1}$). A linear-elastic placeholder under large-deformation kinematics (St Venant–Kirchhoff) has a load maximum at about 42% compression. In designs with top copper thicker than bottom copper and a ceramic of 0.5 mm or less, the DBC edge curls down during the solder-free heat-up and crushed the thin placeholder layers under the DBC corner past this maximum (97.5% compression and an inverted element in a diagnostic run), so 12 of 46 designs attempted with linear placeholders failed at 120–230 °C. The Neo-Hookean placeholder stiffens monotonically in compression; in the verification runs its peak compression was at most 49.7% and its minimum volume ratio at least 0.61.

### 4.2. Solder Creep, Boundary Conditions, Outputs and Mesh Convergence

The Anand parameters of the reference study [1,65] are used: $A = 3501\,\mathrm{s}^{-1}$, $Q/R = 9320\,\mathrm{K}$, $\tilde{s} = 30.2$ MPa, $h_0 = 180000$ MPa, $\xi = 4$, $m = 0.25$, $n = 0.01$, $a = 1.78$ and $s_0 = 21$ MPa. CalculiX has no built-in Anand law, so the steady-state Anand flow rule is fitted per temperature to Norton creep rows at 5 °C intervals. The fit reproduces the Anand strain rate within a factor of 3.8 over the relevant stress range, and transient hardening is neglected.

A 3-2-1 constraint at the bottom corners of the baseplate removes rigid-body motion, following the reference study. Three outputs are extracted for each design. Warpage is defined over the measurement region Ω as

$$W = \max_{\Omega} w - \min_{\Omega} w, \tag{3}$$

where $w$ is the out-of-plane displacement, evaluated at each process stage. The ceramic stress $\sigma_c$ is the maximum over the reflow, housing and molding stages of the signed (tensile-positive) maximum principal stress averaged over a 1.0 × 1.0 × 0.1 mm layer at the copper-edge corner, with the stage of the maximum recorded for each design, and the solder strain $\varepsilon_p$ is the accumulated plastic strain averaged over a 1.0 × 1.0 × 0.1 mm interface layer at the critical corner at the end of molding. The warpage objective is the warpage at the end of molding. Region averages replace peak values because the peaks lie on geometric singularities and do not converge under mesh refinement. Figure 2 shows the three output fields of the nominal design.

Warpage converges within 0.3% on all meshes tested. Over successive refinements, the region-averaged ceramic stress changes by −5.9% and −2.9%, and the solder strain by 0% and +3%. The production mesh has about 23,000 elements and differs from the finest mesh by +0.2% in warpage, −2.9% in solder strain and +9.5% in ceramic stress. A post-registration check on four designs with a finer corner-refined mesh (61,000 elements) gives ceramic-stress changes of −15.85% to +15.06% against the production mesh. The bias is therefore not uniform across designs (spread 30.9 percentage points, against the 3-point rule of the mesh study), and the ceramic-stress values carry a design-dependent discretization error of this size (Supplementary Table S7); warpage changes by at most 0.03% and solder strain by at most 4.82%.

**Sampling mechanism (post-registration).** The ceramic-stress metric selects the integration points that lie within 0.1 mm of the ceramic face. These points sit at fixed fractions (0.211 and 0.789) of each ceramic element layer, whose thickness is the ceramic thickness divided by the number of layers, so the cut at 0.1 mm keeps or drops whole layers of points (Figure 2b and c). On the five designs whose full stress fields were stored, $\sigma_c$ is a step function of the averaging depth, jumping by up to 19% where the depth crosses an integration-point layer (for example 20.9 to 16.9 MPa between 0.09 and 0.10 mm for the 0.3 mm ceramic). In Table S7 the deepest sampled point moves between the production and the finer mesh as follows: D01 0.076 → 0.092 mm (−15.9%); D20 0.091 → 0.089 mm (−1.9%); D37 0.097 → 0.095 mm (−2.0%); D55 0.099 → 0.066 mm (+15.1%). Where it stays put, $\sigma_c$ changes by about 2%; where it moves, by about 15%, in the direction of the through-depth stress gradient. The same rule biases the database itself: on the production mesh a 0.635 mm ceramic is sampled only down to 0.034 mm and a 0.25 mm ceramic down to 0.076 mm, so $\sigma_c$ depends on the ceramic thickness partly through the sampling, not only through the mechanics. This non-physical, step-wise dependence is a plausible contributor to the poor test accuracy of the ceramic-stress surrogates (Section 7). All ceramic-stress results of this paper are therefore reported as indicative.

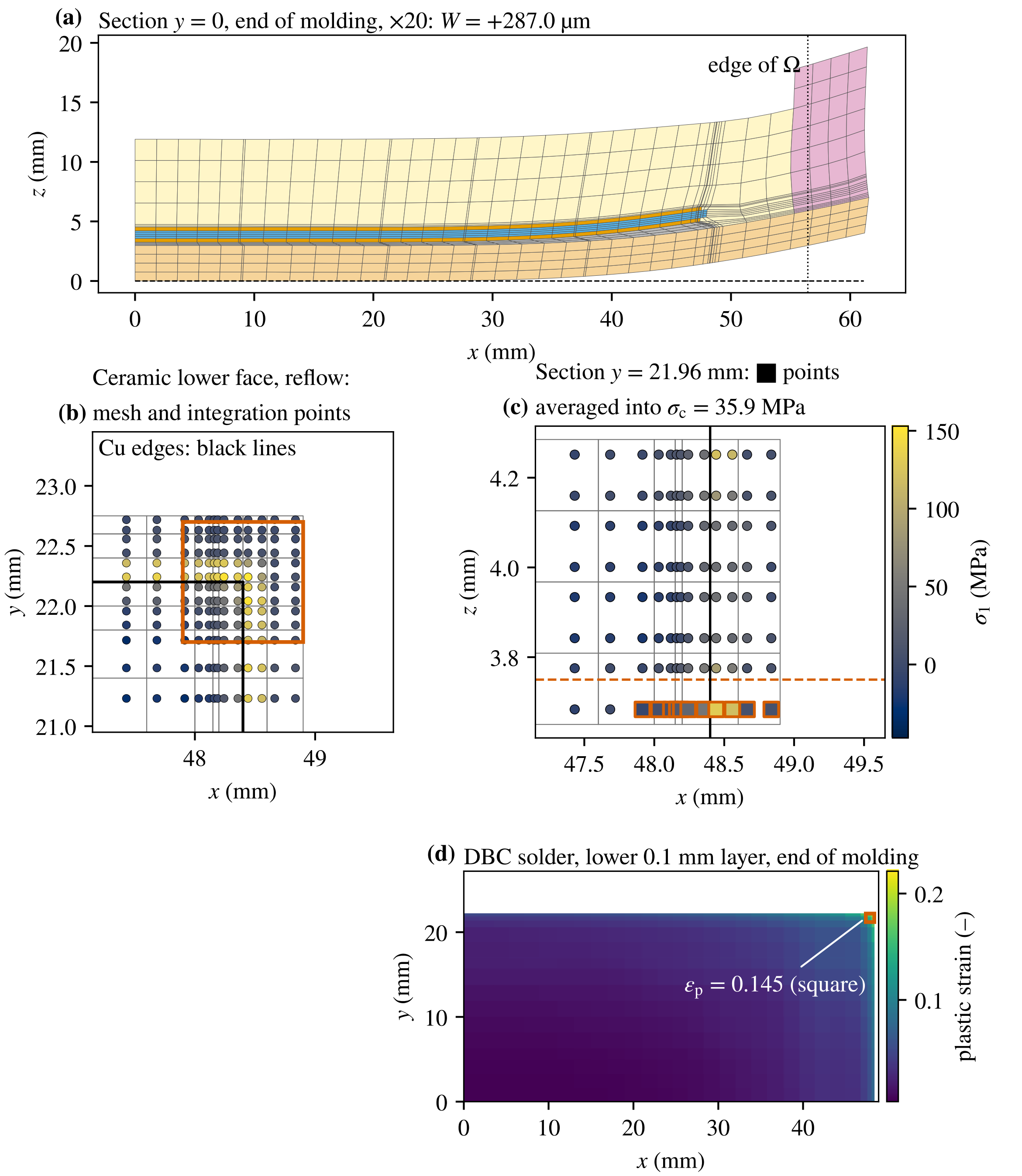

Figure 2. FE output fields of the nominal design (AlN, 0.3/0.635/0.3 mm) from its stored production-mesh run. (a) Section $y = 0$ at the end of molding with the displacements magnified 20 times (colors as in Figure 1; dashed: flat baseplate bottom); $W = +287.0$ µm. (b) Lower ceramic face at the end of reflow, the stage of the $\sigma_c$ maximum: element edges and integration points colored by the maximum principal stress $\sigma_1$; the black lines are the Cu edges and the square is the 1.0 × 1.0 mm averaging region. (c) Section through the corner region: on the production mesh the ceramic has 4 element layers, and only the integration points marked by squares, which lie within the 0.1 mm averaging depth (dashed), enter $\sigma_c = 35.9$ MPa; this sampling makes $\sigma_c$ mesh-sensitive (Section 4.2). (d) Accumulated plastic

strain in the lower 0.1 mm layer of the DBC-to-baseplate solder at the end of molding (element means of the integration points); the square at the solder corner gives $\varepsilon_{p} = 0.145$.

### 4.3. Verification and Validation Against Measured Warpage

The production model, FEA-v5.3 (ring-frame housing, Anand-based creep, start at the second reflow, Neo-Hookean placeholder elements and no fitted parameters), reproduces the measured warpage within 18.2% at reflow, 33.6% at housing and 15.3% at molding (Table 4). The earlier variants in Table 4 simulate the first reflow explicitly and are kept to document how the model was built. Pre-bending (+599.6 μm) is an input rather than a validation point. The measurements are the means reported in [1] from 7, 7 and 3 samples.

Table 4. Measured and predicted warpage (μm) at the three validation stages; absolute errors against the measured mean in parentheses. All variants were fixed before their runs.

| Model variant | Reflow (−376.4) | Housing (−194.8) | Molding (+339.0) |
|---|---|---|---|
| Commercial model of [1], for reference | −355 (5.7%) | −185 (5.0%) | +329 (2.9%) |
| No creep, solid housing block | −377.8 (0.4%) | −379.0 (94.5%) | +116.7 (65.6%) |
| No creep, ring-frame housing | −377.8 (0.4%) | −202.7 (4.1%) | +108.5 (68.0%) |
| Creep, solid housing block | −323.2 (14.1%) | −289.5 (48.6%) | +290.4 (14.3%) |
| Creep, ring-frame housing (FEA-v4-ref) | −323.2 (14.1%) | −130.5 (33.0%) | +279.7 (17.5%) |
| Calibrated (creep ×0.3, 2.5 mm wall) | −333.3 (11.4%) | −158.7 (18.5%) | +254.3 (25.0%) |
| Second-reflow start, linear placeholders (FEA-v5) | −307.6 (18.3%) | −130.0 (33.3%) | +288.3 (15.0%) |
| Second-reflow start, Neo-Hookean placeholders (FEA-v5.3, production) | −307.9 (18.2%) | −129.4 (33.6%) | +287.0 (15.3%) |

**Step changes.** For FEA-v5.3, the change from reflow to housing is +178.5 μm against +181.6 μm measured (1.7% low), so the housing step itself is captured and most of the housing error is the reflow offset carried forward. The change from housing to molding is +416.4 μm against +533.8 μm measured, 22.0% low, and it stays 21.6–23.2% low in every variant tested.

**Housing geometry.** Copper yield strength decreasing with temperature, and shrinkage of a rim adhesive at a nominal 0.192% and at a 3% upper bound, each changed the housing warpage by less than 1 μm. Replacing the solid housing block by a ring frame reduced the housing error from 94.5% to 4.1% without creep, so the housing geometry, not missing physics, drove the housing error.

**Creep.** Without creep, reflow and housing match within 0.4% and 4.1%, but molding is 68% low. With creep, molding improves to 17.5% while reflow degrades to 14.1%. No single creep strength fits all three stages, which points to the steady-state creep simplification: relaxation during fast reflow cooling is overestimated, whereas relaxation and cure effects during the 3 h bake at 150 °C are underestimated.

**Calibration as a negative result.** A pre-registered calibration fitted two unreported quantities on the reflow and housing stages only: a creep-rate multiplier in [0.3, 3] and the frame (bond mode, and wall thickness from 2.5 to 4.5 mm), evaluated on a fixed grid of 30 runs. The optimum lay on the lower bound of both parameters, reached only 11.5% and 18.5% on the calibration stages, and increased the held-out molding error from 17.5% to 25.0%. The calibration was therefore discarded, and no parameter of the production model is fitted.

**Model versions and full rerun.** The placeholder law is not output-neutral near the failing corner of the design space. Relative to FEA-v5 (linear placeholders), FEA-v5.3 changes the baseline design by up to 0.8% (ceramic stress at molding; warpage by at most 0.45% and solder strain by at most 0.07%), a ZTA design with 0.2 mm top copper, 0.3 mm ceramic and 0.1 mm bottom copper by up to 3.9% (solder strain), and a ZTA design with 0.3/0.635/0.2 mm layers by up to 2.3% (solder strain). Because FEA-v5.3 is not equivalent to FEA-v5 within the pre-set 0.5% limit, all 60 designs were rerun with FEA-v5.3, including those that had converged with FEA-v5, with the same design set, train/test split and run order. FEA-v5 results appear only in the supplementary comparison (Table S1). On the 34 designs that converged under both models, FEA-v5.3 changes the warpage by 0.25% on average (at most 0.96%), the ceramic stress by 1.34% (at most 3.70%; more than 2% in 10 designs) and the solder strain by 0.94% (at most 9.47%, in D52, $Si_3N_4$ 0.1/0.32/0.3 mm; more than 2% in 6 designs). The largest changes are not confined to the corner that failed with FEA-v5. Because every optimizer is compared on the same model, the comparisons in Sections 6 and 7 do not depend on its absolute accuracy against the physical module.

## 5. Proposed SSO-Based Surrogate Design Framework

### 5.1. Simplified Swarm Optimization

SSO is used in two roles, to tune the SVR surrogates and to search the design space on those surrogates, and the best designs are then re-simulated by FEA (Figure 3). The return path in Figure 3 is optional: additional FEA cases can be placed where the surrogate is least certain or the designs most promising.

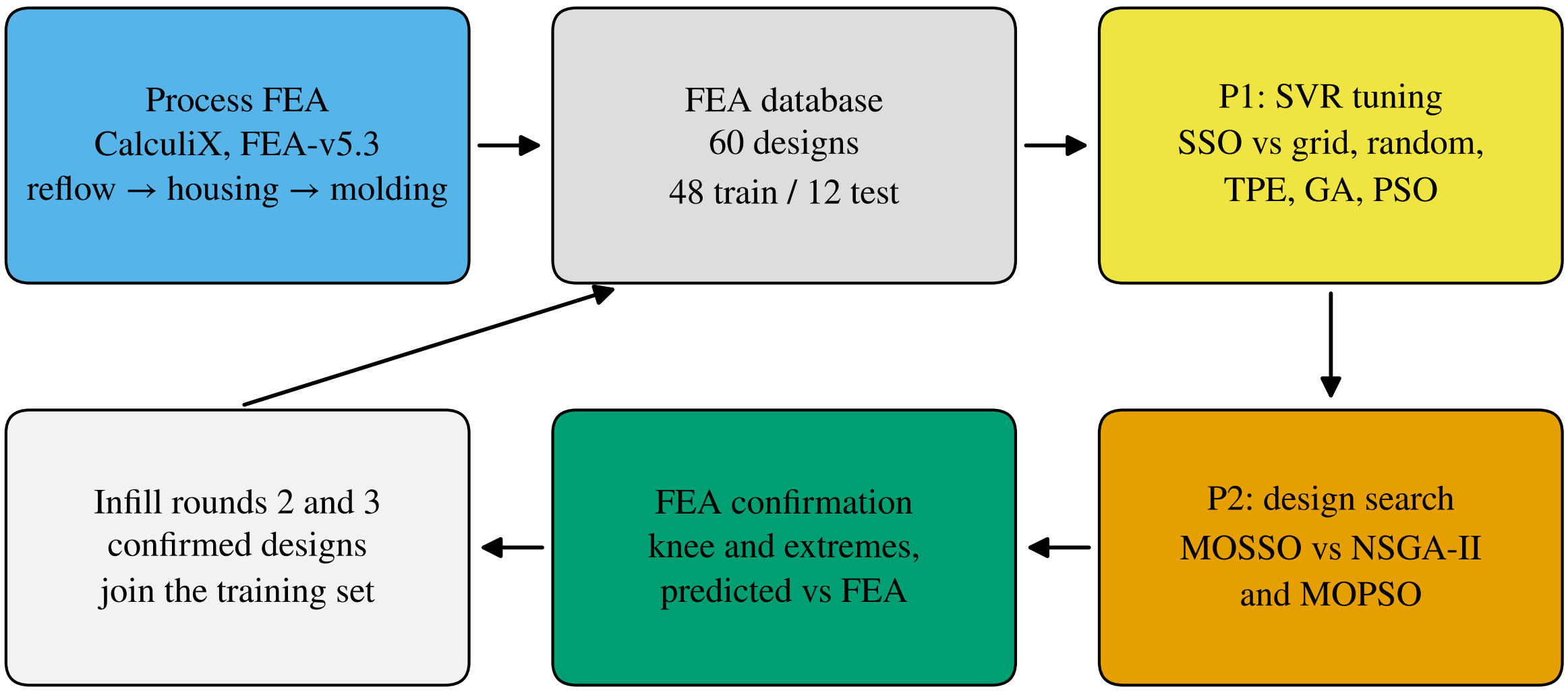

Figure 3. Proposed framework. The process FE model (CalculiX, FEA-v5.3) builds the 60-design database (48 training and 12 test designs). SSO tunes the SVR surrogates (Problem P1; compared with the published grid, random search, TPE, GA and PSO) and searches the design space on them (Problem P2; compared with NSGA-II and MOPSO). The knee and single-objective extreme designs are confirmed by FEA; in the post-registration infill rounds 2 and 3 the confirmed designs join the training set.

For solution $i$ and variable $j$ at iteration $t$, SSO draws $\rho \sim U(0,1)$ and updates the variable by a stepwise rule with thresholds $0 < C_g < C_p < C_w < 1$:

$$x_{ij}^{t+1} = \begin{cases} g_j, & \rho \in [0, C_g), \\ p_{ij}, & \rho \in [C_g, C_p), \\ x_{ij}^{t}, & \rho \in [C_p, C_w), \\ x, & \rho \in [C_w, 1), \end{cases} \tag{4}$$

where $g_j$ is variable $j$ of the global best solution, $p_{ij}$ that of the personal best of solution $i$, and $x$ a random value in the range of variable $j$. Here the random value is drawn uniformly over the full range of the variable, and the categorical material is encoded as a code in [0, 4) mapped by its integer part, the same encoding for every method. The values used are $C_g = 0.4$, $C_p = 0.7$ and $C_w = 0.9$ (global best 0.4, personal best 0.3, current value 0.2, random value 0.1), with a population of 50 for P1 and 100 for P2 and the evaluation budget as the stopping rule.

### 5.2. SSO-Tuned Support Vector Regression Surrogates

SSO solves Problem P1 (Equation (1)) for each of the three outputs, with the SVR trained on the 48 training cases. As in SSO-based hyperparameter tuning of neural networks [56], SSO searches $\log_{10} C$, $\log_{10} \varepsilon$ and $\log_{10} \gamma$ as continuous variables, and the fitness is the five-fold cross-validated $R^2$ on the training set only. The 12 test cases are used once, after tuning, to report test $R^2$. The baselines are the grid of the reference study ($C$ from 1 to 10, and $\varepsilon$ and $\gamma$ from 0.1 to 1, with 10 levels each and a linear or RBF kernel), random search, Bayesian optimization (a multivariate tree-structured Parzen estimator, TPE, as implemented in Optuna [42]), GA and PSO, all with the same budget of 2,000 model fits per run; the grid uses its full 2,000 combinations.

### 5.3. Multi-Objective SSO Design Search and FEA Confirmation

Multi-objective SSO (MOSSO) solves Problem P2 (Equation (2)) on the surrogates. It applies the SSO update with the global best replaced by a leader drawn from an external archive. The archive holds every non-dominated solution found; above 100 members, the member with the smallest crowding distance is removed repeatedly. For each solution and generation, the leader is the winner of a binary tournament on crowding distance between two random archive members, and a personal best is replaced when the new solution dominates it, or with probability 0.5 when neither dominates. NSGA-II [45], in the pymoo implementation [50], and MOPSO [46] serve as baselines with the same 20,000 surrogate evaluations and a population of 100, and a weighted single-objective version (mean of the three normalized objectives) is compared with GA and PSO. Post-registration, MOPSO was corrected to the published algorithm of [46], and every multi-objective method reports the same external archive (Section 7). From the run with the median hypervolume of each method, the knee point (smallest distance to the origin of the normalized objective space) and the three single-objective extremes are re-simulated with FEA-v5.3; only FEA-confirmed values are reported as design results. Post-registration, the confirmed designs are added to the training set and the surrogates are re-tuned (confirmation-driven infill, Section 7).

## 6. Experimental Design

**Evaluation questions.** The experiments answer three questions fixed before any optimization run. Q1: does SSO tuning give a higher test $R^2$ than the published grid and the other tuners at an equal number of model fits? Q2: does multi-objective SSO reach a larger hypervolume than NSGA-II and MOPSO at an equal number of surrogate evaluations? Q3: do the surrogate-optimal designs keep their predicted advantage when re-simulated by FEA?

**FEA database.** The design levels follow the reference study: top copper of 0.1, 0.2 or 0.3 mm, ceramic of 0.25, 0.3, 0.32, 0.5 or 0.635 mm, bottom copper of 0.1, 0.2 or 0.3 mm, and four ceramics, 180 combinations in total. The reference study ran 60 of them without listing which, so a balanced, seeded subset (seed 20260928) is selected here. Each ceramic material has 15 designs in which every pairing of ceramic thickness with top-copper thickness, and of ceramic thickness with bottom-copper thickness, appears exactly once. Overall, each copper level appears in 20 designs, each ceramic thickness in 12, and each top and bottom copper pair 6 or 7 times. A 48/12 train/test split is stratified by material, with three test designs per material covering all five ceramic thicknesses and each copper level four times. Each case runs the full process with FEA-v5.3 in 2.9 h on average (2.2–3.8 h) on 4 threads, and six cases run in parallel on a workstation with an Intel Core i7-14700 processor and 125 GB of memory. A case that fails to converge is retried once with FEA-v5.3.1, which changes only the incrementation controls of the first process step and was verified on the baseline design to change no output by more than 0.15%. No design is replaced, and the train/test split is not redrawn. All 60 designs converged at the first attempt, so no retry was needed and the 48/12 split is complete; the database took 174 case-hours (697 core-hours). Confirmation-driven infill cases (post-registration) are described in Section 7.

**Protocols and statistics.** Table 5 lists the comparison protocol. All methods use the same training data, feasibility bounds, cost definitions, hardware and stopping criteria, and every method receives an equal, documented tuning effort on the training data only.

Table 5. Comparison protocol.

| Item | Setting |
|---|---|
| Independent runs per method | 30, with fixed and reported seeds |
| Budget for SVR tuning | 2,000 model fits per run and output |
| Budget for design search | 20,000 surrogate evaluations per run, population 100 |
| Surrogate metrics | test $R^2$ and root-mean-square error on the 12 held-out cases |
| Multi-objective metrics | hypervolume with a common reference point, inverted generational distance, spread |
| Summary statistics | mean, median, standard deviation and 95% confidence interval |
| Paired tests | Wilcoxon signed-rank test, exact $p$-values, Holm correction, effect sizes |
| Ranking | Friedman test across methods |
| Confirmation | FEA-v5.3 runs of the knee and the three extremes of the median-hypervolume run of each method |

**Reproducibility.** The FE model is frozen by file hash, the design list and all seeds are released, and every reported number is produced by a logged script. The database differs from that of the reference study in the 60 designs, the region-averaged outputs and the solver, so surrogate accuracies are compared on this database with the grid of the reference study rather than with its reported $R^2$ values.

**Pre-registration and later changes.** The protocol of Sections 5 and 6 (bounds, parameters, budgets, seeds, statistics and the confirmation rule) was frozen with its code hash before any P1 or P2 run. One deviation was made before launch: the $C$ upper bound was widened to $10^4$. After the pre-registered runs, four items were added and are labeled post-registration wherever they appear: (i) a correction of the P2 result reporting so that every multi-objective method reports the same external archive, (ii) a correction of MOPSO to the published algorithm, (iii) two rounds of confirmation-driven infill, and (iv) diagnostics on standard benchmark problems. The pre-registered results are reported alongside.

## 7. Results

All numbers in this section are produced by logged scripts from the frozen database (SHA-256 998643a5…) and the frozen code. Pre-registered results are reported first; post-registration corrections and infill rounds are labeled as such.

**FEA database.** All 60 designs converged with FEA-v5.3 at the first attempt; no design needed the retry and none failed, so the 48 training and 12 test designs are complete. Across the 60 designs, the warpage at the end of molding ranges from +257.6 to +452.8 μm (mean 350.7, standard deviation 47.2), the ceramic stress from 10.6 to 39.0 MPa (mean 19.6, standard deviation 5.6) and the solder strain from 0.053 to 0.218 (mean 0.098, standard deviation 0.040). The ceramic stress peaks at the reflow stage in 58 designs and at the housing stage in 2 (AlN 0.1/0.635/0.3, $Si_3N_4$ 0.1/0.635/0.2 mm), where the housing value exceeds the reflow value by at most 0.7 MPa. By material, $Si_3N_4$ gives the lowest mean warpage (310.1 μm) with a mean ceramic stress of 22.9 MPa and solder strain of 0.136, whereas $Al_2O_3$ gives the highest mean warpage (384.1 μm) with 16.8 MPa and 0.074; AlN gives 332.6 μm, 21.4 MPa and 0.108; ZTA gives 376.2 μm, 17.5 MPa and 0.076. Warpage and solder strain are strongly anti-correlated (Pearson $r = -0.90$, Spearman $\rho = -0.95$), and the ceramic stress is weakly correlated with both ($r = -0.31$ with warpage and 0.26 with solder strain), so the three objectives conflict and P2 is a genuine multi-objective problem. Ceramic thickness has the largest main effect on warpage (level means from 392.4 μm at 0.25 mm to 305.2 μm at 0.635 mm) and on solder strain (0.071 to 0.138), followed by the ceramic material, and top-copper thickness has the largest main effect on the ceramic stress (15.3 MPa at 0.1 mm to 23.6 MPa at 0.3 mm).

**Surrogate accuracy (Q1, pre-registered).** Table 6 reports the pre-registered P1 results (Q1). Figure 4 shows the predicted against the FEA values of the shared surrogates used in P2 for the training and test designs. On the training-only fitness, every stochastic tuner beats the published grid for all three outputs (median cross-validated $R^2$ 0.986, 0.811 and 0.934 for SSO against 0.951, 0.762 and 0.908 for the grid). On the 12 test designs this advantage holds only for warpage (test $R^2$ 0.9934 against 0.9634). For the ceramic stress and the solder strain the grid generalizes better (test $R^2$ 0.380 against tuner medians 0.306–0.315; 0.889 against 0.801–0.859). Among the five tuners, test accuracy is statistically indistinguishable: after Holm correction no comparison of SSO with another tuner is significant for the ceramic stress or the solder strain, and for warpage SSO is below random search by a median of 0.0001 ($p_{\text{Holm}} = 0.045$); the Friedman tests on test $R^2$ give $p = 0.085$, 0.15 and 0.045. The cross-validated ranking does not transfer: GA, PSO and TPE rank above SSO on cross-validated $R^2$ for every output (Friedman $p \leq 3.2 \times 10^{-10}$), but not on test $R^2$. The answer to Q1 is therefore negative: SSO tuning does not give a higher test $R^2$ than the other tuners, and continuous tuning beats the published grid on the test set only for warpage. The final $C$ lies within 10% of the widened bound in 6/150 warpage, 0/150 ceramic stress, 33/150 solder strain runs (all tuners). The ceramic-stress results are indicative, because the ceramic-stress metric is mesh-sensitive at the production resolution (Section 4.2, Supplementary S8).

Table 6. Pre-registered P1: test $R^2$ on the 12 test designs (median over 30 runs) and, in parentheses, the five-fold cross-validated $R^2$ on the 48 training designs. The published grid is one deterministic run.

| Output | Grid [1] | Random | TPE | GA | PSO | SSO |
|---|---|---|---|---|---|---|
| Final warpage | 0.9634 (0.9511) | 0.9936 (0.9854) | 0.9935 (0.9857) | 0.9935 (0.9858) | 0.9935 (0.9858) | 0.9934 (0.9855) |
| Ceramic stress | 0.3799 (0.7625) | 0.3062 (0.8115) | 0.3135 (0.8138) | 0.3120 (0.8140) | 0.3144 (0.8140) | 0.3149 (0.8113) |
| Solder strain | 0.8891 (0.9078) | 0.8492 (0.9340) | 0.8007 (0.9355) | 0.8044 (0.9348) | 0.8595 (0.9348) | 0.8305 (0.9338) |

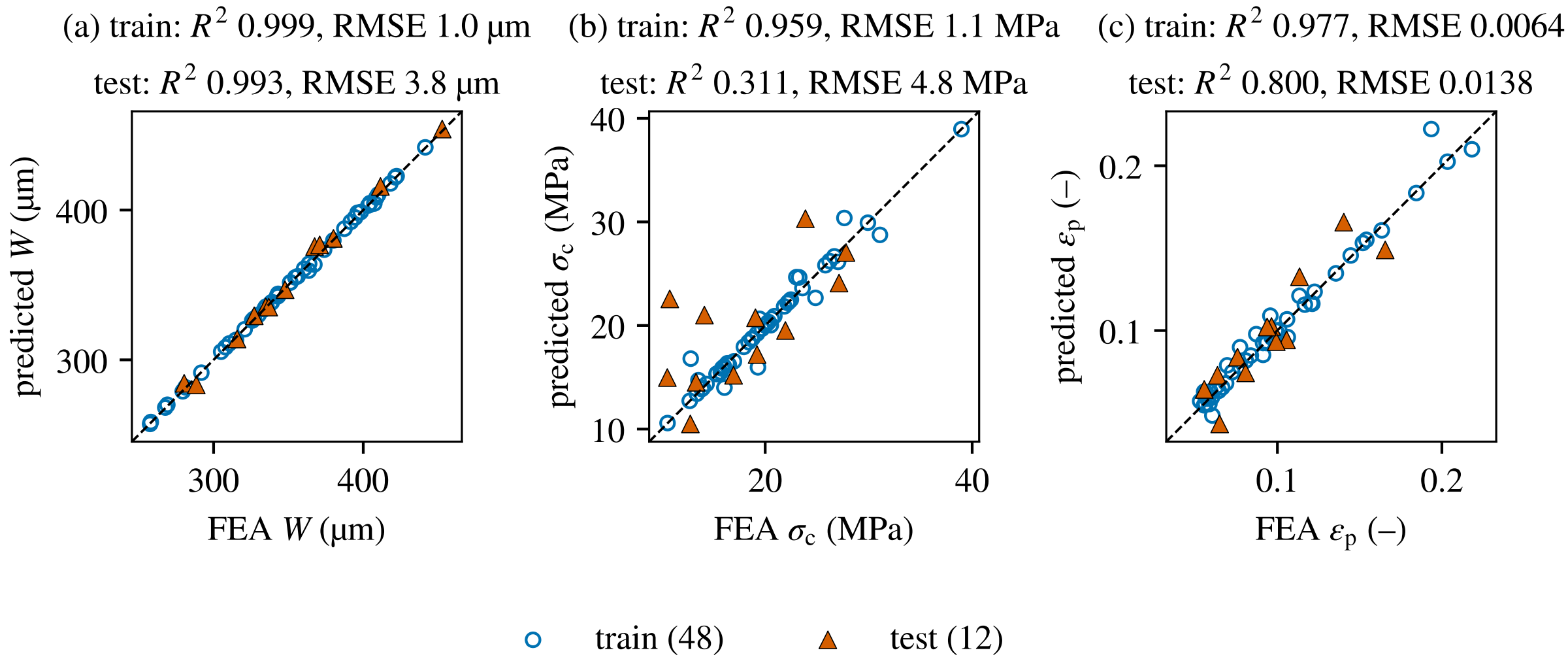

Figure 4. Predicted against FEA values of the shared P2 surrogates (for each output the SSO-tuned SVR of the P1 run with the highest cross-validated $R^2$, refitted on the 48 training designs): (a) warpage $W$, (b) ceramic stress $\sigma_c$, (c) solder strain $\varepsilon_p$. Circles: training designs; triangles: test designs; dashed: equality line. The panel titles give $R^2$ and the RMSE of each set; the test values equal those of Table 6 for these runs. The $\sigma_c$ values are indicative (Section 4.2).

**Design optimization (Q2).** As pre-registered, each method reported its own result set: MOSSO its external archive, NSGA-II its final population and MOPSO its repository. Under that reporting MOSSO reached the largest hypervolume (median 1.1094 against 1.1045 for NSGA-II and 0.5607 for MOPSO; $p_{\text{Holm}} = 1.9 \times 10^{-8}$ and $3.7 \times 10^{-9}$). Two post-registration diagnostics showed that this comparison was not symmetric (Supplementary S3). Giving NSGA-II the same external archive raises its hypervolume to MOSSO's level on the same seeds, and the MOPSO implementation departed from the published algorithm in its mutation schedule, its mutation sampling and the tie-breaking of its archive. Table 7 reports the corrected comparison (post-registration correction): every method reports the same external archive, and MOPSO follows [46]. MOSSO and NSGA-II are then comparable: their hypervolumes do not differ ($p_{\mathrm{Holm}}=0.4$), MOSSO has a small edge in inverted generational distance ($p_{\mathrm{Holm}}=0.043$, effect size +0.42) and the spreads do not differ ($p_{\mathrm{Holm}}=0.67$). The corrected MOPSO trails both ($p_{\mathrm{Holm}}=3.7\times10^{-9}$ against MOSSO). In the weighted single-objective comparison GA and PSO reach a lower weighted objective than SSO by about 0.02% ($p_{\mathrm{Holm}}=3.7\times10^{-9}$). Figure 5 shows the median-hypervolume fronts. Because the ceramic stress is one of the three objectives, the fronts are indicative in that dimension (Section 4.2).

Table 7. P2, round 1: hypervolume (HV), inverted generational distance (IGD) and spread, median [IQR] over 30 runs. Pre-registered reporting: each method's own result set (MOPSO as pre-registered). Corrected reporting (post-registration): the same external archive for every method, MOPSO corrected to [46]. $p_{\mathrm{Holm}}$: Wilcoxon signed-rank test against MOSSO on the corrected HV.

| **Method** | **HV, pre-registered reporting** | **HV, corrected** | **IGD, corrected** | **Spread, corrected** | $p_{\mathrm{Holm}}$ **(HV)** |
|---|---|---|---|---|---|
| MOSSO | 1.1094 | 1.1094 [1.1082–1.1114] | 0.0280 [0.0271–0.0286] | 0.432 | — |
| NSGA-II | 1.1045 | 1.1092 [1.1074–1.1108] | 0.0287 [0.0277–0.0295] | 0.409 | 0.4 |
| MOPSO (pre-registered / corrected) | 0.5607 | 1.0867 [1.0836–1.0922] | 0.0534 [0.0469–0.0615] | 0.480 | $3.7\times10^{-9}$ |

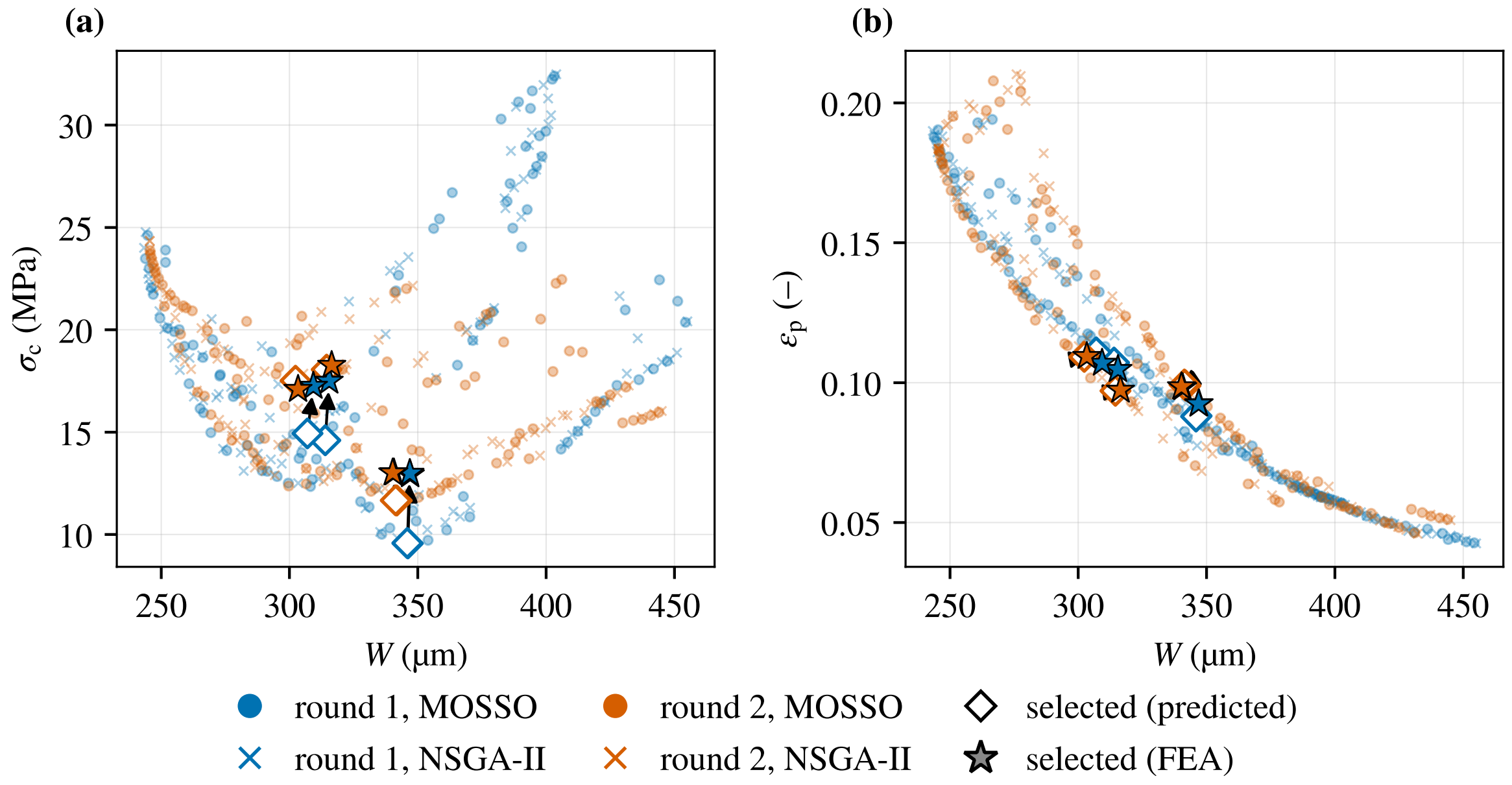

Figure 5. Median-hypervolume fronts of MOSSO (circles) and NSGA-II (crosses) with equal external archives: predicted (a) warpage $W$ against ceramic stress $\sigma_c$ and (b) $W$ against solder

strain $\varepsilon_p$, for round 1 (blue) and the post-registration infill round 2 (vermilion). For the knee and best-$\sigma_c$ designs of each method, the open diamonds are the surrogate predictions and the stars the FEA values of the same designs; the arrows from prediction to FEA show the optimism of the $\sigma_c$ surrogate at these designs. The $\sigma_c$ values are indicative (Section 4.2).

**FEA confirmation (Q3, pre-registered).** The pre-registered rule selected 15 designs (13 distinct), all of which converged with FEA-v5.3. Warpage is predicted within 1.1% (mean absolute error 0.5%), but the ceramic stress is under-predicted by 16.5% and 13.4% at the MOSSO and NSGA-II knees and by 26.3% and 26.0% at their best-$\sigma_c$ designs (mean absolute error over the 13 designs 16.4%). The best-$\varepsilon_p$ designs extrapolate below the database minimum and miss by up to 33%; their FEA results are dominated by 6 database designs. Of the 13 designs, 9 are not dominated by any of the 60 database designs and none dominates the nominal design. The best confirmed warpage, 245.3 µm, is 4.8% below the best database design (257.6 µm, D58), whereas the best confirmed ceramic stress, 12.92 MPa, is above the database minimum (10.55 MPa, D22). The answer to Q3 is therefore mixed: the predicted advantage survives FEA for warpage but not for the solder strain, nor for the ceramic stress as measured by the production-resolution metric; since that metric is mesh-sensitive (Section 4.2), the ceramic-stress part of this answer is indicative. Figure 6 compares predicted and FEA values for every confirmed design.

Table 8. FEA confirmation, round 1: FEA-v5.3 results of the 13 distinct selected designs and prediction error (predicted − FEA)/FEA; designs selected by several methods show the error of the first selection. Dominated: by the 60 database designs.

| **Design** | **Selected by** | **FEA $W$ (µm)** | **FEA $\sigma_c$ (MPa)** | **FEA $\varepsilon_p$** | **Error $W$** | **Error $\sigma_c$** | **Error $\varepsilon_p$** | **Dominated** |
|---|---|---|---|---|---|---|---|---|
| $Si_3N_4$ 0.100/0.635/0.300 | NSGA-II best $W$ | 245.3 | 24.08 | 0.1836 | −0.8% | −0.3% | +3.5% | no |
| $Si_3N_4$ 0.101/0.624/0.299 | MOSSO best $W$ | 246.6 | 23.80 | 0.1816 | −1.1% | −1.3% | +3.4% | no |
| $Si_3N_4$ 0.109/0.440/0.228 | MOPSO best $W$ | 273.7 | 18.07 | 0.1578 | −0.4% | −9.7% | −2.5% | no |
| $Si_3N_4$ 0.109/0.402/0.218 | MOPSO best $\sigma_c$ | 280.9 | 18.21 | 0.1487 | +0.3% | −13.5% | −2.9% | no |
| $Si_3N_4$ 0.109/0.374/0.223 | MOPSO best $\varepsilon_p$; MOPSO knee | 287.3 | 18.62 | 0.1378 | +0.7% | −13.6% | −2.1% | no |
| AlN 0.100/0.415/0.271 | NSGA-II knee | 309.3 | 17.25 | 0.1070 | −0.7% | −13.4% | +3.6% | no |
| AlN 0.101/0.385/0.251 | MOSSO knee | 315.5 | 17.49 | 0.1048 | −0.5% | −16.5% | +2.3% | no |
| ZTA 0.100/0.445/0.180 | SSO (weighted) optimum | 343.8 | 12.96 | 0.0941 | −0.4% | −25.0% | −5.3% | 1 |
| ZTA 0.100/0.444/0.180 | GA (weighted) optimum; PSO (weighted) optimum | 343.8 | 12.96 | 0.0941 | −0.4% | −25.0% | −5.3% | 1 |
| ZTA 0.101/0.415/0.159 | MOSSO best $\sigma_c$ | 346.9 | 12.99 | 0.0924 | −0.3% | −26.3% | −4.8% | no |
| ZTA 0.100/0.412/0.153 | NSGA-II best $\sigma_c$ | 347.0 | 12.92 | 0.0925 | −0.3% | −26.0% | −4.7% | no |
| $Al_2O_3$ 0.232/0.251/0.299 | MOSSO best $\varepsilon_p$ | 453.9 | 16.84 | 0.0635 | +0.1% | +20.9% | −32.6% | 6 |
| $Al_2O_3$ 0.232/0.250/0.300 | NSGA-II best $\varepsilon_p$ | 454.6 | 16.87 | 0.0636 | +0.1% | +21.1% | −33.1% | 6 |

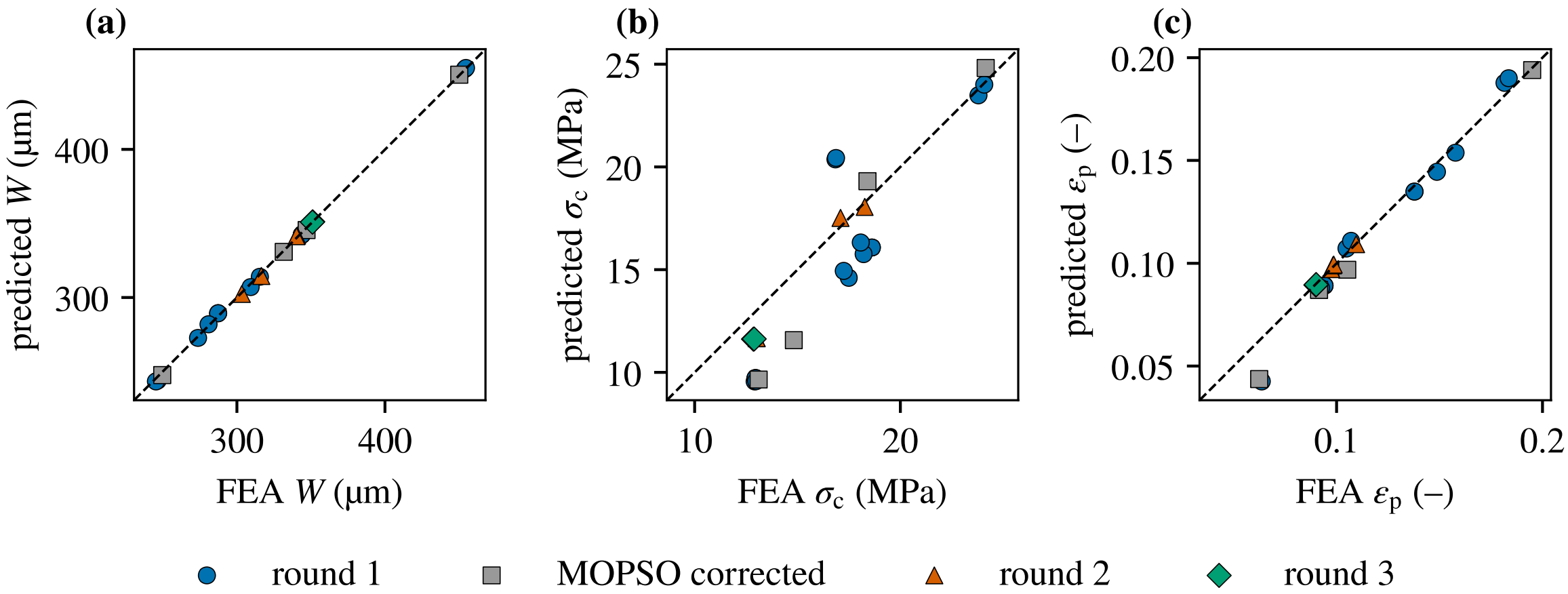

Figure 6. Predicted against FEA values of (a) warpage $W$, (b) ceramic stress $\sigma_c$ and (c) solder strain $\varepsilon_p$ for every confirmed design: round 1, the corrected MOPSO (post-registration) and the infill rounds 2 and 3 (post-registration); dashed: equality line. The $\sigma_c$ values are indicative (Section 4.2).

**Confirmation-driven infill (post-registration).** The confirmed designs were then added to the training set (post-registration), keeping the test set unchanged. Round 2 added the 13 round-1 designs (61 training designs), re-tuned the three surrogates with the SSO arm of the pre-registered protocol and repeated P2 with equal archives. The median test $R^2$ of the ceramic-stress surrogate rose from 0.315 to 0.397 (cross-validated $R^2$ 0.811 to 0.847). On four designs that neither round had trained on (the corrected-MOPSO confirmations), the mean absolute error fell from 14.0% to 6.7% for the ceramic stress and from 10.8% to 2.0% for the solder strain. The ceramic-stress error of the knees fell to −1.2% (MOSSO) and +2.5% (NSGA-II), and that of the best-$\sigma_c$ designs to −10.3% and −10.4%. A third, narrow round added the eight designs confirmed in round 2 and in the MOPSO correction (69 training designs) and confirmed only the best-$\sigma_c$ designs; their ceramic-stress errors are −9.8% (MOSSO) and −9.6% (NSGA-II). Table 9 and Figure 7 summarize the rounds. Infill corrected most of the error at the knees, but the surrogate remains optimistic where the ceramic stress is lowest, and the lowest confirmed ceramic stress of any design (12.86 MPa) is still above the database minimum (10.55 MPa). With equal archives, MOSSO's hypervolume edge over NSGA-II is not significant in round 1 ($p_{\text{Holm}} = 0.4$) and significant in round 2 ($p_{\text{Holm}} = 0.013$); in round 3 it is not ($p_{\text{Holm}} = 0.46$), and its inverted generational distance is lower in every round; the two methods select near-identical designs. All ceramic-stress errors in this paragraph refer to the production-resolution metric and are indicative (Section 4.2).

Table 9. Confirmation-driven infill (post-registration): ceramic-stress ($\sigma_c$) surrogate accuracy and $\sigma_c$ prediction error (predicted − FEA)/FEA of the confirmed designs of the median-hypervolume runs. Round 3 confirms only the best-$\sigma_c$ designs. Out-of-sample: the four corrected-MOPSO designs, trained on in round 3. Indicative: the $\sigma_c$ metric is mesh-sensitive at the production resolution (Section 4.2, Supplementary S8).

| **Quantity** | **Round 1 (48 training)** | **Round 2 (61 training)** | **Round 3 (69 training)** |
| --- | --- | --- | --- |
| $\sigma_c$ test $R^2$, median of 30 SSO runs | 0.315 | 0.397 | 0.490 |
| $\sigma_c$ cross-validated $R^2$, median | 0.811 | 0.847 | 0.843 |
| $\sigma_c$ out-of-sample MAPE (4 designs) | 14.0% | 6.7% | — (in training) |
| MOSSO knee, $\sigma_c$ error | −16.5% | −1.2% | not selected |
| NSGA-II knee, $\sigma_c$ error | −13.4% | +2.5% | not selected |
| MOSSO best $\sigma_c$, $\sigma_c$ error (FEA $\sigma_c$) | −26.3% (12.99) | −10.3% (13.01) | −9.8% (12.89) |
| NSGA-II best $\sigma_c$, $\sigma_c$ error (FEA $\sigma_c$) | −26.0% (12.92) | −10.4% (13.02) | −9.6% (12.86) |
| MOSSO vs NSGA-II, HV $p_{\text{Holm}}$ (equal archives) | 0.4 | 0.013 | 0.46 |

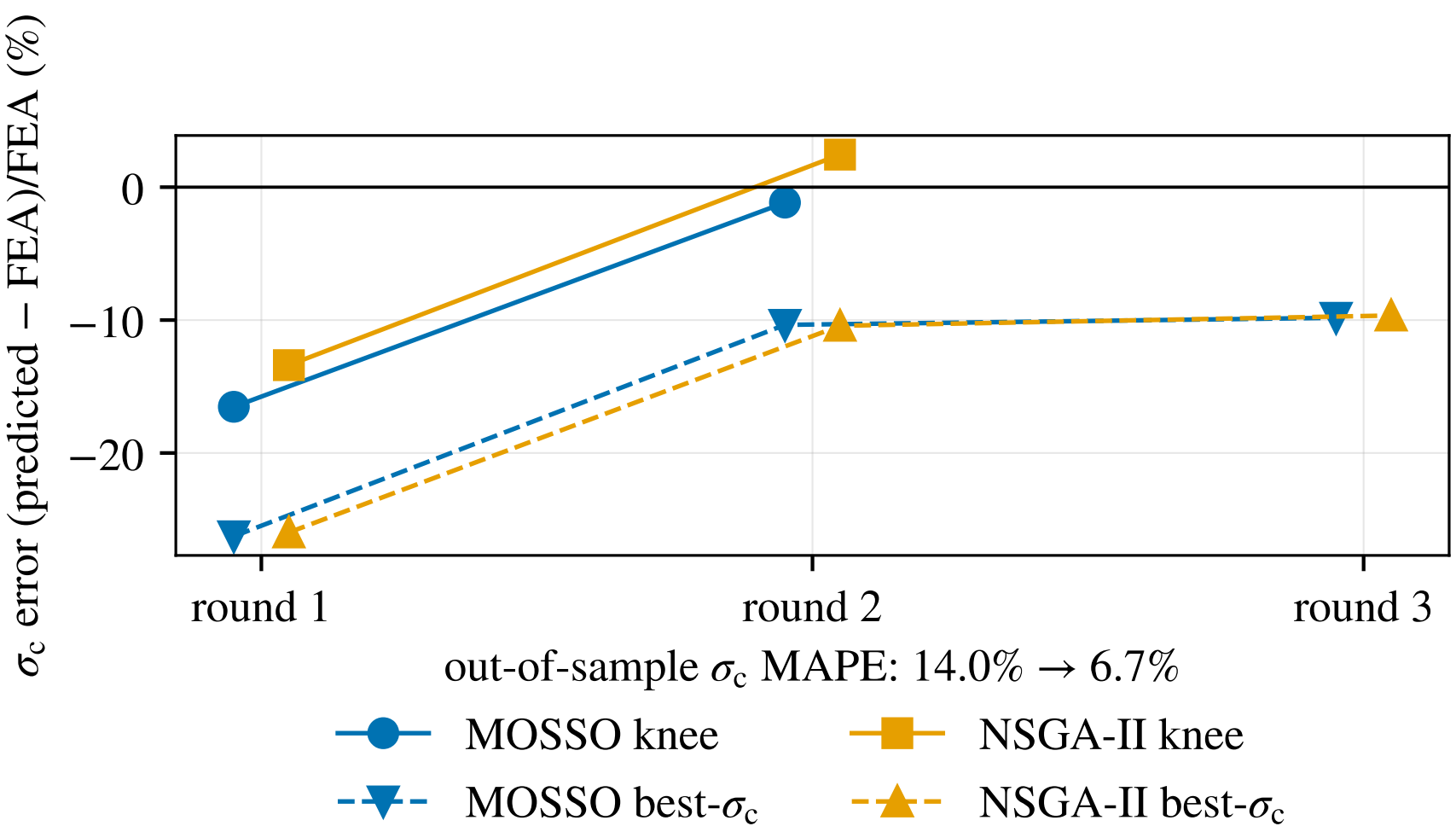

Figure 7. Ceramic-stress prediction error, (predicted − FEA)/FEA, of the knee and best-$\sigma_c$ designs of the median-hypervolume MOSSO and NSGA-II runs, by round (rounds 2 and 3: post-registration infill; round 3 confirms only the best-$\sigma_c$ designs). The legend title gives the out-of-sample $\sigma_c$ MAPE of rounds 1 and 2. Indicative (Section 4.2).

**Sensitivity.** Halving the creep tolerance (CETOL $2.5 \times 10^{-4}$ instead of $5 \times 10^{-4}$) on the nominal design changes the warpage by +0.18%, the ceramic stress by −0.007% and the solder strain by +0.11% (at most 0.27% at any stage), so the production tolerance is adequate. In a post-registration sensitivity study (Supplementary Table S6), changing $C_g$ or $C_p$ by ±0.1 or lowering $C_w$ to 0.8 leaves the hypervolume and the inverted generational distance of MOSSO unchanged (all $p_{\text{Holm}} \geq 0.53$); only $C_w = 1.0$, which switches the random reset off, is worse (hypervolume 1.0570 against 1.0721, $p_{\text{Holm}} = 0.012$). The population size was not varied. On four database designs re-run on a finer mesh with 61,000 instead of 23,000 elements (post-registration, Supplementary Table S7), the warpage changes by at most 0.03%, the ceramic stress by −15.85% to +15.06% and the solder strain by +1.67% to +4.82%. Warpage is insensitive to the mesh, but the ceramic-stress metric is not: at some designs its mesh-induced change is as large as, or larger than, the surrogate's residual ceramic-stress error at the design-space bounds after infill (−9.6% to −10.4% at the best-$\sigma_c$ designs of rounds 2 and 3). At the production resolution, the residual ceramic-stress confirmation error therefore cannot be separated from the discretization error of the ceramic-stress metric itself. The changes differ in sign between designs, which is explained by the integration-point sampling of the 0.1 mm averaging layer (Section 4.2). One further design (D14) stopped with a solver ordering error and is listed in Table S7. A depth-consistent definition of the ceramic stress removes the sign-flipping artifact but leaves a one-sided finer-mesh bias of 7.5–10.3% on three of the four designs (+0.7% on the fourth; post-registration, Supplementary S8).

## 8. Discussion

**Optimizers exploit surrogate error at the bounds.** The central finding, indicative for the ceramic stress, is that the optimizers exploit surrogate error at the edges of the design space, a known failure mode of optimization on response surfaces [35]. Of the front members, 48.5% (MOSSO) and 36.3% (NSGA-II) have the top copper at its lower bound, and 82.3% and 68.5% have some thickness within 1% of a bound (Supplementary Table S4; Figure 8), where the database is thinnest. The largest ceramic-stress errors occur exactly there: the designs selected for the lowest ceramic stress are the most optimistic, by about a quarter in round 1. The test set shows the same weakness from the other side: its three largest ceramic-stress residuals (D18 +11.8 MPa, D39 +6.9 MPa, D09 +6.4 MPa) are thin-ceramic designs with thick copper on both faces, a corner represented by a single training design (Supplementary S5). Confirmation-driven infill is the direct remedy: it moves training data to where the optimizer goes, and two rounds removed most of the error at the knees. It did not produce a design that beats the database on ceramic stress, which suggests, with the caveat below, that the reported fronts are limited by the surrogate in that corner rather than by the optimizers. The mesh check (Supplementary Table S7) adds a caveat: the ceramic-stress metric itself changes by −15.9% to +15.1% with the mesh at some designs, so part of the residual ceramic-stress error may be discretization error of the metric rather than surrogate error.

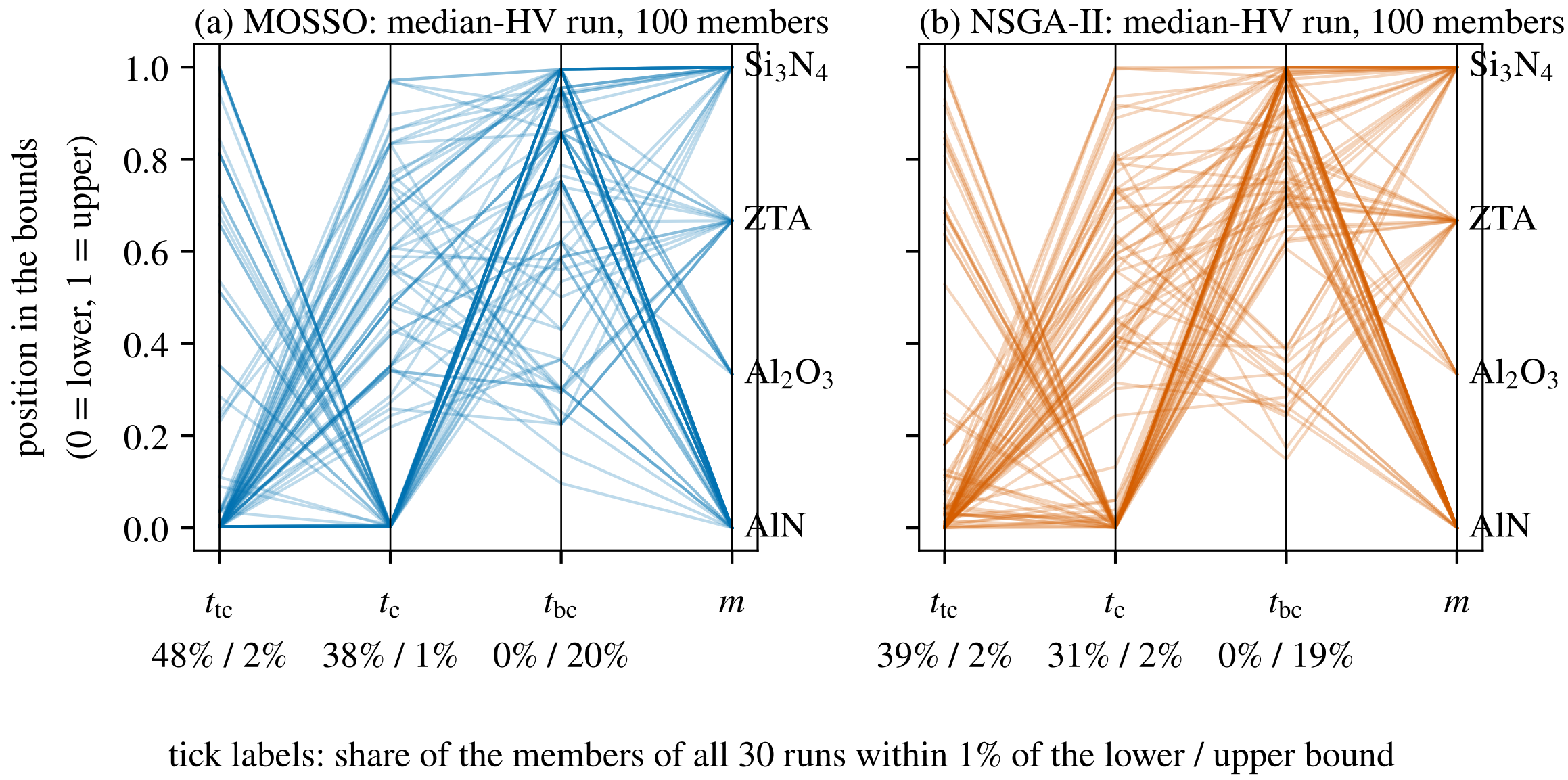

Figure 8. Front members of the median-hypervolume runs of (a) MOSSO and (b) NSGA-II in round 1 (equal external archives) as parallel coordinates: each line is one design, with $t_{tc}$, $t_c$

and $t_{bc}$ scaled to their bounds (0 = lower, 1 = upper) and the ceramic material $m$ on the last axis. The tick labels give, over all 30 runs, the share of members within 1% of the range at the lower / upper bound; the same rule applied to the pre-registered reporting gives Supplementary Table S4.

**Tuning and generalization.** With 48 training designs, a better cross-validated fit does not mean a better test fit [43]: the tuners that maximize the five-fold cross-validated $R^2$ most (GA, PSO, TPE) are not better on the test set, and the coarse published grid, with the lowest cross-validated $R^2$, generalizes best for the ceramic stress and the solder strain. A plausible reading is that its small $C$ values regularize the model; the study was not designed to test this. For surrogate tuning on databases of this size, the choice of tuner matters less than the coverage of the training data.

**MOSSO and NSGA-II.** With equal result reporting the two are comparable on P2, with a small, consistent MOSSO edge in inverted generational distance and a hypervolume edge in the infill rounds. On standard problems the picture differs (post-registration benchmarks, Supplementary S2): on ZDT1 with 4, 10 and 30 variables MOSSO reaches a hypervolume of 0.867, 0.852 and 0.565 against 0.871, 0.871 and 0.868 for NSGA-II, and on DTLZ2 0.692 against 0.699. The per-variable random reset of SSO slows the convergence of many tail variables, so MOSSO suits low-dimensional design problems such as P2 (four variables) better than high-dimensional ones.

**MOPSO.** Corrected to the published algorithm [46] (mutation schedule and sampling, and random tie-breaking in the archive; post-registration), MOPSO still trails on the benchmarks (hypervolume 0.311 on ZDT1 and 0.178 on DTLZ2). With the published 30 divisions per objective, three objectives give 27,000 hypercubes for a 100-member repository, so nearly every member occupies its own hypercube and the grid cannot rank crowding; the retention rule then becomes random deletion.

**Model fidelity.** The production model is within 15% to 34% of the measured warpage without fitting, against 2.9% to 5.7% for the commercial model of [1]. The gap has identified causes. The housing geometry, chip layout, mounting holes and pre-bend shape are not reported and had to be assumed, and the open-source solver lacks the full Anand law and a cure-dependent viscoelastic EMC model. The housing-to-molding step is 22% to 23% low in every variant, which points to the EMC cure behavior described in [21].

**Scope of the claims.** All optimizers run on the same surrogates and the same FE model, so their ranking does not depend on the absolute accuracy of the model. The optimized designs are predictions for the modeled module; transferring them to the physical module requires the missing geometry or experimental confirmation. This study is a simulation-level validation and does not claim hardware-level validation of the optimized designs.

**Cost.** Each FEA case takes 2.9 h on average with FEA-v5.3 on 4 threads, whereas a surrogate evaluation takes milliseconds. The 60-case database took 697 core-hours. All 453 P1 runs took 11.0 core-hours and the 180 corrected P2 runs 0.32 core-hours, against 23 confirmation FEA cases of about 3 h each.

## 9. Conclusions

An open-source process FE model of an IGBT-DBC module, validated within 18.2–33.6% of measured warpage without fitted parameters, was combined with SSO-tuned surrogates, multi-objective design search, FEA confirmation and confirmation-driven infill. The pre-registered comparisons were negative or neutral for SSO: the tuners tie on test accuracy, and with equal result reporting multi-objective SSO and NSGA-II give comparable fronts. The pipeline's main result is diagnostic and, for the ceramic stress, indicative: FEA confirmation exposed optimistic ceramic-stress predictions at the design-space bounds, where the optimizers concentrate their designs, and infill reduced that error from about a quarter to about a tenth at the best-$\sigma_c$ designs without yielding a design that improves on the database in ceramic stress. A mesh check qualifies this finding: the ceramic-stress metric changes by up to 16% between meshes at some designs, and even a depth-consistent definition keeps a one-sided bias of up to 10%, so a mesh-converged stress metric is a prerequisite for firm ceramic-stress conclusions. Future work concerns adaptive sampling of the bounds before optimization, transient creep and cure modeling of the EMC, and a second packaging case.

## Data and Code Availability

The code and the FEA database supporting this study will be deposited in Zenodo and made publicly available upon acceptance (DOI to be added upon acceptance). The FE model runs in CalculiX 2.21 with the SPOOLES solver on meshes generated by gmsh 4.15.2, driven by Python 3.12.3 with NumPy 2.5.3 and SciPy 1.18.1; the surrogates and optimizers use scikit-learn 1.9.1, Optuna 5.0.0 and pymoo 0.6.2. All runs were made under Linux (kernel 6.8) on an Intel Core i7-14700 processor with 125 GB of memory; the design list, all random seeds, the pre-registration and the file hashes of the frozen model and database are part of the deposit.

## Author Contributions

Wei-Chang Yeh: Conceptualization, Methodology, Software, Validation, Formal analysis, Investigation, Data curation, Writing – original draft, Writing – review & editing, Visualization, Project administration, Funding acquisition.

## Acknowledgments

This work was supported in part by the National Science and Technology Council (NSTC), Taiwan, R.O.C., under grant nos. NSTC 114-2221-E-007-123-MY3 and NSTC 113-2221-E-007-117-MY3.

## Declaration of Interests

The author declares no known competing financial interests or personal relationships that could have appeared to influence the work reported in this paper.

## Declaration of Generative AI and AI-Assisted Technologies in the Writing Process

During the preparation of this work the author used a generative AI assistant to help draft and copy-edit text and to assist with computational scripting and the generation of figures and result tables. After using this tool, the author reviewed and edited the content as needed and takes full responsibility for the content of the publication.

**Supplementary Material**

Table S1. FEA-v5 (linear placeholders) and FEA-v5.3 (Neo-Hookean placeholders, production) on the 34 designs converged under both: warpage $W$ at the end of molding (μm), ceramic stress $\sigma_c$ as the maximum over stages (MPa) and solder strain $\varepsilon_p$ at the end of molding; Δ is the change of FEA-v5.3 relative to FEA-v5 (%). Design: ceramic, then top copper/ceramic/bottom copper thickness (mm).

| **Case** | **Design** | $W$ **v5** | $W$ **v5.3** | $\Delta W$ | $\sigma_c$ **v5** | $\sigma_c$ **v5.3** | $\Delta \sigma_c$ | $\varepsilon_p$ **v5** | $\varepsilon_p$ **v5.3** | $\Delta \varepsilon_p$ |
|---|---|---|---|---|---|---|---|---|---|---|
| D01 | AlN 0.1/0.25/0.1 | 348.8 | 347.8 | −0.30 | 13.83 | 13.32 | −3.66 | 0.0964 | 0.0964 | +0.03 |
| D05 | AlN 0.2/0.3/0.2 | 352.1 | 351.3 | −0.22 | 19.74 | 19.45 | −1.44 | 0.0911 | 0.0913 | +0.25 |
| D07 | AlN 0.1/0.32/0.2 | 330.7 | 330.1 | −0.18 | 14.14 | 14.34 | +1.40 | 0.0988 | 0.0990 | +0.26 |
| D09 | AlN 0.3/0.32/0.3 | 368.1 | 367.5 | −0.14 | 24.08 | 23.89 | −0.79 | 0.0805 | 0.0807 | +0.19 |
| D10 | AlN 0.1/0.5/0.1 | 293.3 | 291.9 | −0.50 | 14.17 | 13.87 | −2.08 | 0.1637 | 0.1634 | −0.17 |
| D11 | AlN 0.2/0.5/0.3 | 306.0 | 305.1 | −0.31 | 22.22 | 22.32 | +0.45 | 0.1214 | 0.1204 | −0.80 |
| D13 | AlN 0.1/0.635/0.3 | 269.3 | 267.8 | −0.59 | 19.30 | 20.01 | +3.70 | 0.1489 | 0.1446 | −2.88 |
| D14 | AlN 0.2/0.635/0.2 | 281.9 | 280.4 | −0.55 | 27.37 | 27.14 | −0.84 | 0.1657 | 0.1655 | −0.10 |
| D16 | $Al_2O_3$ 0.1/0.25/0.2 | 418.1 | 418.2 | +0.01 | 16.51 | 16.35 | −0.95 | 0.0587 | 0.0587 | +0.01 |
| D18 | $Al_2O_3$ 0.3/0.25/0.3 | 453.0 | 452.8 | −0.03 | 10.89 | 10.78 | −0.97 | 0.0650 | 0.0649 | −0.09 |
| D19 | $Al_2O_3$ 0.1/0.3/0.3 | 421.1 | 421.5 | +0.10 | 16.20 | 15.85 | −2.18 | 0.0554 | 0.0555 | +0.09 |
| D22 | $Al_2O_3$ 0.1/0.32/0.1 | 371.4 | 370.9 | −0.13 | 10.77 | 10.55 | −2.05 | 0.0757 | 0.0758 | +0.07 |
| D23 | $Al_2O_3$ 0.2/0.32/0.3 | 422.2 | 422.3 | +0.02 | 17.14 | 16.99 | −0.87 | 0.0561 | 0.0561 | −0.03 |
| D25 | $Al_2O_3$ 0.1/0.5/0.2 | 343.8 | 343.2 | −0.18 | 12.63 | 12.72 | +0.73 | 0.0938 | 0.0936 | −0.14 |
| D28 | $Al_2O_3$ 0.1/0.635/0.3 | 343.4 | 342.9 | −0.15 | 15.27 | 15.31 | +0.24 | 0.0956 | 0.0913 | −4.42 |
| D29 | $Al_2O_3$ 0.2/0.635/0.2 | 335.5 | 334.8 | −0.20 | 22.09 | 21.96 | −0.57 | 0.1055 | 0.1056 | +0.09 |
| D31 | ZTA 0.1/0.25/0.2 | 407.5 | 407.4 | −0.00 | 16.19 | 16.07 | −0.76 | 0.0563 | 0.0563 | −0.05 |
| D32 | ZTA 0.2/0.25/0.3 | 441.4 | 441.5 | +0.02 | 19.47 | 19.22 | −1.24 | 0.0605 | 0.0605 | −0.08 |
| D34 | ZTA 0.1/0.3/0.3 | 409.7 | 410.1 | +0.08 | 16.05 | 15.72 | −2.06 | 0.0533 | 0.0531 | −0.27 |
| D35 | ZTA 0.2/0.3/0.1 | 380.4 | 380.1 | −0.08 | 16.56 | 16.94 | +2.30 | 0.0654 | 0.0636 | −2.78 |
| D37 | ZTA 0.1/0.32/0.1 | 364.4 | 363.8 | −0.15 | 10.81 | 10.58 | −2.15 | 0.0808 | 0.0808 | +0.06 |
| D38 | ZTA 0.2/0.32/0.2 | 387.8 | 387.5 | −0.09 | 13.70 | 13.55 | −1.08 | 0.0664 | 0.0665 | +0.19 |
| D39 | ZTA 0.3/0.32/0.3 | 411.7 | 411.5 | −0.03 | 14.24 | 14.14 | −0.74 | 0.0554 | 0.0556 | +0.33 |
| D40 | ZTA 0.1/0.5/0.2 | 337.5 | 336.9 | −0.17 | 12.67 | 12.77 | +0.83 | 0.0993 | 0.0993 | +0.01 |
| D43 | ZTA 0.1/0.635/0.3 | 336.0 | 335.3 | −0.19 | 15.28 | 15.35 | +0.44 | 0.1010 | 0.0972 | −3.85 |
| D45 | ZTA 0.3/0.635/0.2 | 335.2 | 334.8 | −0.12 | 26.27 | 26.23 | −0.17 | 0.0987 | 0.0964 | −2.33 |
| D49 | $Si_3N_4$ 0.1/0.3/0.2 | 311.6 | 310.6 | −0.31 | 15.68 | 16.03 | +2.22 | 0.1159 | 0.1167 | +0.72 |
| D51 | $Si_3N_4$ 0.3/0.3/0.3 | 351.9 | 351.2 | −0.20 | 27.93 | 27.65 | −0.99 | 0.0931 | 0.0934 | +0.27 |
| D52 | $Si_3N_4$ 0.1/0.32/0.3 | 315.7 | 315.7 | +0.01 | 19.22 | 19.06 | −0.82 | 0.1036 | 0.0938 | −9.47 |
| D53 | $Si_3N_4$ 0.2/0.32/0.2 | 322.1 | 320.8 | −0.39 | 20.55 | 20.18 | −1.80 | 0.1209 | 0.1213 | +0.26 |
| D55 | $Si_3N_4$ 0.1/0.5/0.1 | 281.6 | 279.4 | −0.78 | 14.36 | 13.98 | −2.66 | 0.2043 | 0.2034 | −0.42 |
| D56 | $Si_3N_4$ 0.2/0.5/0.3 | 270.7 | 269.1 | −0.59 | 23.13 | 23.29 | +0.69 | 0.1524 | 0.1519 | −0.32 |
| D59 | $Si_3N_4$ 0.2/0.635/0.1 | 284.0 | 281.3 | −0.96 | 22.45 | 22.17 | −1.23 | 0.1949 | 0.1935 | −0.71 |
| D60 | $Si_3N_4$ 0.3/0.635/0.3 | 259.9 | 258.1 | −0.70 | 39.19 | 38.96 | −0.60 | 0.1846 | 0.1844 | −0.12 |

Mean $|\Delta|$: $W$ 0.25%, $\sigma_c$ 1.34%, $\varepsilon_p$ 0.94%; maximum $|\Delta|$: $W$ 0.96%, $\sigma_c$ 3.70%, $\varepsilon_p$ 9.47%. Values from the frozen database (database_v53.csv) at full precision. The stage of the $\sigma_c$ maximum is the same under both models for every design.

Table S2. Post-registration benchmark diagnostics (P2 settings, 10 runs each, hypervolume median, reference point 1.1). MOPSO: as pre-registered and corrected to [46].

| **Problem** | **True front** | **MOSSO** | **NSGA-II** | **MOPSO (pre-reg.)** | **MOPSO (corrected)** |
|---|---|---|---|---|---|
| ZDT1, 30 variables | 0.8714 | 0.5648 | 0.8679 | 0.2192 | 0.3105 |
| DTLZ2, 12 variables, 3 objectives | 0.7568 | 0.6921 | 0.6992 | 0.1437 | 0.1780 |
| ZDT1, 4 variables | 0.8714 | 0.8674 | 0.8709 | — | — |
| ZDT1, 10 variables | 0.8714 | 0.8520 | 0.8708 | — | — |

S3. P2 reporting symmetry (post-registration diagnostic). On seeds 0–9, giving NSGA-II the same external archive as MOSSO raises its median hypervolume from 1.1045 to 1.1094, against 1.1095 for MOSSO. The pre-registered P2 result (Table 7, first column) is kept on record; the corrected reporting is used for all later comparisons.

Table S4. Bound activity of the final front members of the 30 P2 runs (round 1): share of members within 1% of a bound.

| Method | Members | Top Cu at lower bound | Ceramic at lower bound | Bottom Cu at upper bound | Any thickness at a bound |
|---|---|---|---|---|---|
| MOSSO | 3000 | 48.5% | 38.0% | 20.1% | 82.3% |
| NSGA-II | 3000 | 36.3% | 29.3% | 18.3% | 68.5% |
| MOPSO (as pre-registered) | 3000 | 0.0% | 0.2% | 0.0% | 0.2% |

S5. Ceramic-stress coverage gap (post-registration diagnostic). Of the four designs with 0.3 mm copper on both faces and a ceramic of 0.32 mm or less, three (D09, D18, D39) are test designs and one (D51, $Si_3N_4$) is a training design; the three carry the largest test residuals. Exploratory variants with the same cross-validation protocol give test $R^2$ 0.311 (pre-registered features), 0.413 ($\log \sigma_c$), 0.492 (ordinal material code) and 0.212 (material properties); none closes the gap, and the pre-registered model is kept.

Table S6. SSO parameter sensitivity (post-registration): MOSSO on the round-3 P2 problem, 10 runs per setting, one threshold changed at a time by ±0.1. HV and IGD median; $p_{\text{Holm}}$: Wilcoxon signed-rank test against the base setting (0.4, 0.7, 0.9), Holm-corrected over the six variants; $r_{\text{rb}} > 0$ means the variant is better.

| Setting | ($C_g$, $C_p$, $C_w$) | HV | IGD | HV $p_{\text{Holm}}$ ($r_{\text{rb}}$) | IGD $p_{\text{Holm}}$ ($r_{\text{rb}}$) |
|---|---|---|---|---|---|
| base | (0.4, 0.7, 0.9) | 1.0721 | 0.0283 | — | — |
| Cg−0.1 | (0.3, 0.7, 0.9) | 1.0730 | 0.0294 | 1 (+0.13) | 1 (−0.31) |
| Cg+0.1 | (0.5, 0.7, 0.9) | 1.0704 | 0.0289 | 1 (−0.16) | 1 (−0.16) |
| Cp−0.1 | (0.4, 0.6, 0.9) | 1.0683 | 0.0286 | 1 (−0.27) | 1 (−0.02) |
| Cp+0.1 | (0.4, 0.8, 0.9) | 1.0722 | 0.0287 | 1 (+0.16) | 1 (−0.27) |
| Cw−0.1 | (0.4, 0.7, 0.8) | 1.0664 | 0.0293 | 0.53 (−0.60) | 1 (−0.05) |
| Cw+0.1 | (0.4, 0.7, 1.0) | 1.0570 | 0.0334 | 0.012 (−1.00) | 0.012 (−1.00) |

Table S7. Mesh-bias check (post-registration): five database designs re-run on the finer corner-refined mesh (61k instead of 23k elements); $\Delta = (\text{fine} - \text{production}) / \text{production}$. Design rule: one per ceramic thickness, materials cycling, the most symmetric copper.

| Design | Material, $t_{\text{tc}}/t_{\text{c}}/t_{\text{bc}}$ (mm) | $\Delta W$ | $\Delta \sigma_{\text{c}}$ | $\Delta \varepsilon_{\text{p}}$ |
|---|---|---|---|---|
| D01 | AlN 0.1/0.25/0.1 | +0.02% | −15.85% | +1.81% |
| D20 | $Al_2O_3$ 0.2/0.3/0.2 | −0.01% | −1.88% | +2.53% |
| D37 | ZTA 0.1/0.32/0.1 | −0.03% | −1.99% | +1.67% |
| D55 | $Si_3N_4$ 0.1/0.5/0.1 | −0.00% | +15.06% | +4.82% |
| D14 | AlN 0.2/0.635/0.2 | failed: SPOOLES ordering error (GPart_makeYCmap) at step 10 on the 747k-equation system; not a convergence failure; not retried because the pre-approved retry targets step 1 only. | | |

Completed designs: 4 of 5. $\Delta\sigma_c$ ranges from −15.85% to +15.06% (spread 30.91 percentage points); $\Delta W$ and $\Delta\varepsilon_p$ stay within 0.03% and 4.82%.

S8. Singularity-avoiding ceramic-stress definitions (post-registration). Four designs were re-run to the end of reflow on the production and the finer mesh with the full stress field kept. Three definitions fixed in advance were evaluated: A, a depth-consistent average over exactly the 1.0 × 1.0 × 0.1 mm corner volume (stress interpolated through the depth between integration points); B, A without the 0.25 mm around the corner vertex; C, the stress at a fixed point 0.5 mm inside the corner, 0.05 mm below the face. The largest mesh-induced change is 15.9% for the original metric, 10.3% for A, 19.0% for B and 57.1% for C; none meets the ±3% criterion. A depth-consistent definition removes the sign-flipping artifact but leaves a one-sided finer-mesh bias of 7.5–10.3% on three of the four designs (+0.7% on the fourth). Definition C lies in a compressive, low-magnitude region. The convergence of the finer mesh itself is not established here. Figure S1 shows the change for each design and definition.

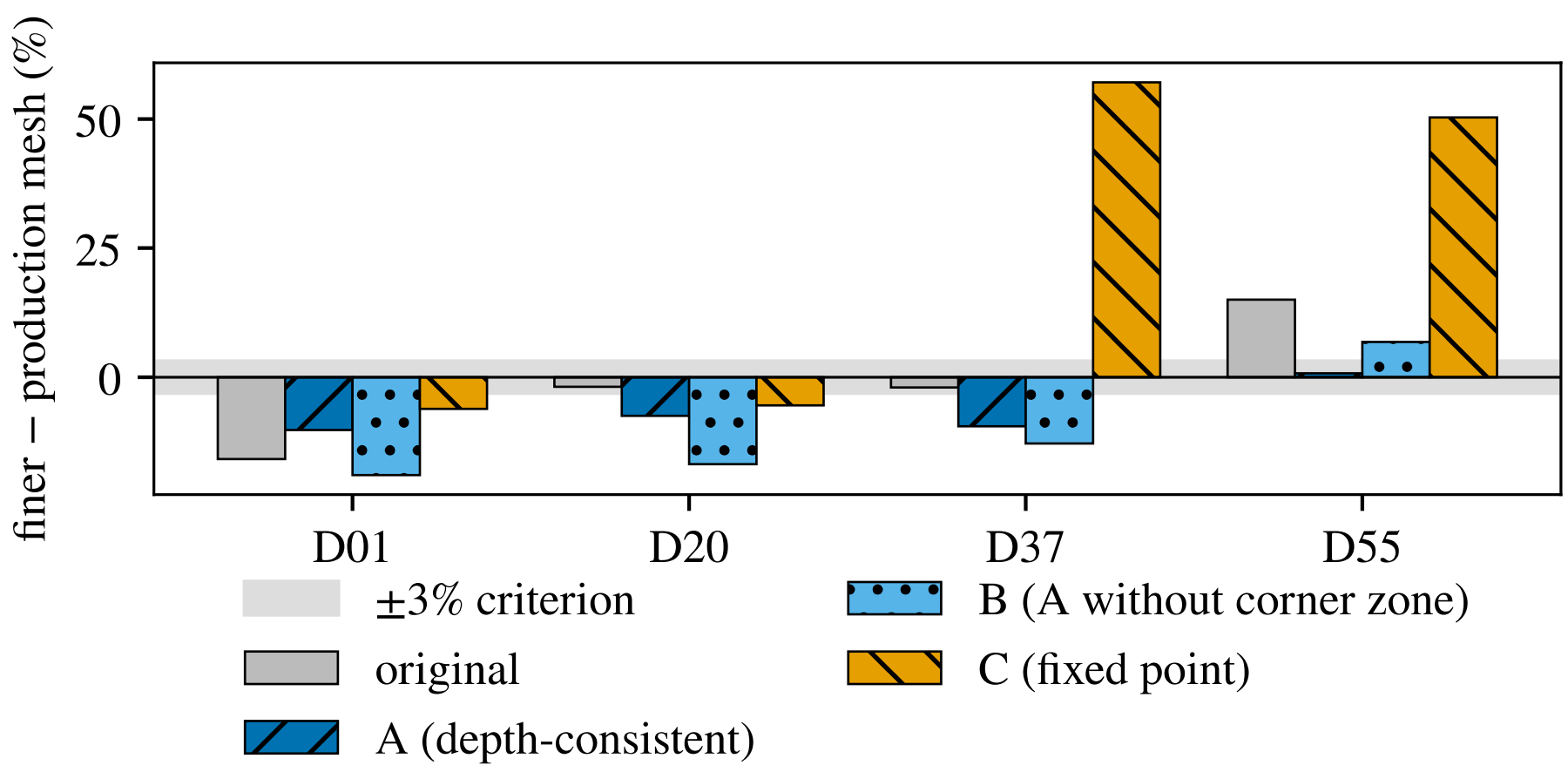

Figure S1. Change of the ceramic stress between the finer corner-refined mesh and the production mesh, $(\text{finer} - \text{production})/\text{production}$, for the four designs of S8 and the original definition and definitions A, B and C; the gray band is the ±3% criterion fixed in advance (post-registration).